\documentclass[letterpaper, 10 pt, conference]{ieeeconf}  % Comment this line out
\IEEEoverridecommandlockouts                              % This command is only
\usepackage{graphicx}
\usepackage{amsmath}
\usepackage{cite}
\usepackage{url}

\usepackage{times}
\usepackage{fancyhdr,graphicx,amsmath,amssymb}
\usepackage[ruled,vlined]{algorithm2e}
\include{pythonlisting}

\usepackage[font=footnotesize,belowskip=-0.5pt,aboveskip=0pt]{caption}
\usepackage{bigstrut}
\usepackage[utf8]{inputenc}
\usepackage{array}
\usepackage{multirow}
\usepackage{tabularx}
\usepackage{bigstrut}
\usepackage{makecell}
\usepackage{tabu}
\usepackage{bm}
\usepackage{caption}

\usepackage{amsmath}          
\usepackage{graphics}
\usepackage{graphicx}
\usepackage{float}
\usepackage{caption}
\usepackage{subcaption}
\usepackage{dblfloatfix}
\usepackage{calc}
\usepackage{url}
\usepackage{textcomp}
\usepackage{balance}
\usepackage{siunitx}
\usepackage{mdwlist}
\usepackage{comment}

\usepackage{gensymb}

\usepackage{array}
\usepackage{multirow}
\usepackage{tabularx}
\usepackage{booktabs}

\usepackage{bigstrut}
\usepackage{fixltx2e}

\usepackage{color}
\usepackage{hhline}

\newcommand{\revision}[1]{\textcolor{black}{#1}}

\usepackage[capitalize]{cleveref}

\title{\LARGE \bf
A Multi-Chamber Smart Suction Cup \\for Adaptive Gripping and Haptic Exploration
}

\author{Tae Myung Huh$^{1}$, Kate Sanders$^{2}$, Michael Danielczuk$^{2}$, Monica Li$^{1}$, Yunliang Chen$^{2}$,\\ Ken Goldberg$^{2}$, Hannah S. Stuart$^{1}$% <-this % stops a space
\thanks{This paper has supplementary downloadable material available at https://youtu.be/-eTlSLk-9jA, provided by the authors. This includes movie clips of sliding haptic exploration, FTIR test setup, and image process of FTIR video data.}%
\thanks{$^{1}$T. M. Huh, M. Li and H. S. Stuart are with the Embodied Dexterity Group, Dept. of Mechanical Engineering, University of California Berkeley, Berkeley, CA, USA. {\tt\small thuh@berkeley.edu}}%
\thanks{$^{2}$K.Sanders, M. Danielczuk, Y. Chen, and K. Goldberg are with AUTOLAB, Dept. of Electrical Engineering and Computer Science, University of California Berkeley, Berkeley, CA, USA.}%
}

\begin{document}

\maketitle

\thispagestyle{empty}
\pagestyle{empty}

\global\csname @topnum\endcsname 0
\global\csname @botnum\endcsname 0

%%%%%%%%%%%%%%%%%%%%%%%%%%%%%%%%%%%%%%%%%%%%%%%%%%%%%%%%%%%%%%%%%%%%%%%%%%%%%%%%
\begin{abstract}

We present a novel robot end-effector for gripping and haptic exploration. Tactile sensing through suction flow monitoring is achieved with a new suction cup design that contains multiple chambers for air flow. Each chamber connects with its own remote pressure transducer, which enables both absolute and differential pressure measures between chambers. By changing the overall vacuum applied to this smart suction cup, it can perform different functions such as gentle haptic exploration (low pressure) and monitoring breaks in the seal during strong astrictive gripping (high pressure). Haptic exploration of surfaces through sliding and palpation % when the suction pump pressure is reduced. }%In this mode, it can detect sealing properties as the suction cup slides across a surface 
can guide the selection of suction grasp locations and help to identify the local surface geometry. During suction gripping, a trained LSTM network can localize breaks in the suction seal between four quadrants with up to 97\% accuracy and detects breaks in the suction seal early enough to avoid total grasp failure. %, and informs an estimate of grip wrench limits. %Conveniently, electronics can be located remotely from the suction cup, allowing it to remain small, flexible and resilient.
%This inexpensive technology is most immediately intended for industrial  settings where suction gripping is already common. 

\end{abstract}

%%%%%%%%%%%%%%%%%%%%%%%%%%%%%%%%%%%%%%%%%%%%%%%%%%%%%%%%%%%%%%%%%%%%%%%%%%%%%%%%

% You can make your own Tex to make your paragraph or sections.
\section{Introduction}
\label{sec:Intro}

Vacuum grippers are widely used in industry to handle objects. They perform astrictive grasping or, in other words, they apply attractive forces to object surfaces through suction pressure. The unicontact suction cup has the advantage of simple operation and enables the handling of a wide range of items, including those that are delicate, large or not accessible by a jaw gripper %. For these reasons, many participants in the Amazon picking challenge utilized suction grippers
\cite{correll2016analysis}.

One major challenge in suction grasping is how to plan a contact location. Examples of planning methods include the heuristic search for a surface normal\cite{morrison2018cartman} and neural network training of grasp affordance using binary success labels\cite{zeng2018robotic}. Wan et al. use CAD model meshes to plan a grasp resisting gravitational wrench\cite{wan2020planning}, and Dex-Net 3.0 learns the best suction contact pose from a point cloud considering both suction seal formation and gravitational wrench resistance\cite{mahler2018dex}. These methods rely on RGB or depth sensors, which may not perceive fine details critical to suction success, e.g., texture, rugosity, porosity, etc. Vision can also become occluded in cluttered environments. %he gripper may work in a cluttered environment where vision is easily occluded. Therefore, 
%Tactile sensors hold the potential to observe this missing information through haptic exploration. 
We investigate how \revision{tactile} sensing can be incorporated into the suction mechanism to monitor local contact geometry through haptic exploration. %and control of suction gripping. %For robust suction grasping in practice, the robotic system may need tactile sensors to monitor surface textures and local contact geometry. 

Another challenge arises during forceful manual manipulation. %Another challenge is maintaining suction grasping while manipulating objects. 
In industry, robotic speed is desired for time efficiency, however the inertial force induced by motion can cause suction grasp failure. Pham et al. use time-optimal path generation bounded by contact stability constraints to generate critically fast arm trajectories during pick-and-place \cite{Pham2019}. Cheng et al. demonstrate an optimal control approach with a single suction gripper to reorient object by extrinsic dexterity, utilizing external contacts
from the table \cite{cheng2019manipulation}. Both methods utilize known inertial properties of the gripped object. These types of dynamic and forceful maneuvers could be adaptively achieved with the addition of suction cup tactile sensing, especially for objects with properties that \revision{are not known \textit{a priori} and} might compromise suction seal. %on the feedback of measured vacuum seal state from the proposed tactile sensor. 

%%%%%
\begin{figure}[tbp!]
\centering
	%\vspace{-10pt}
	%\begin{subfigure}[h]{0.5\textwidth}
	%\centering
	\includegraphics[width=0.9\linewidth]{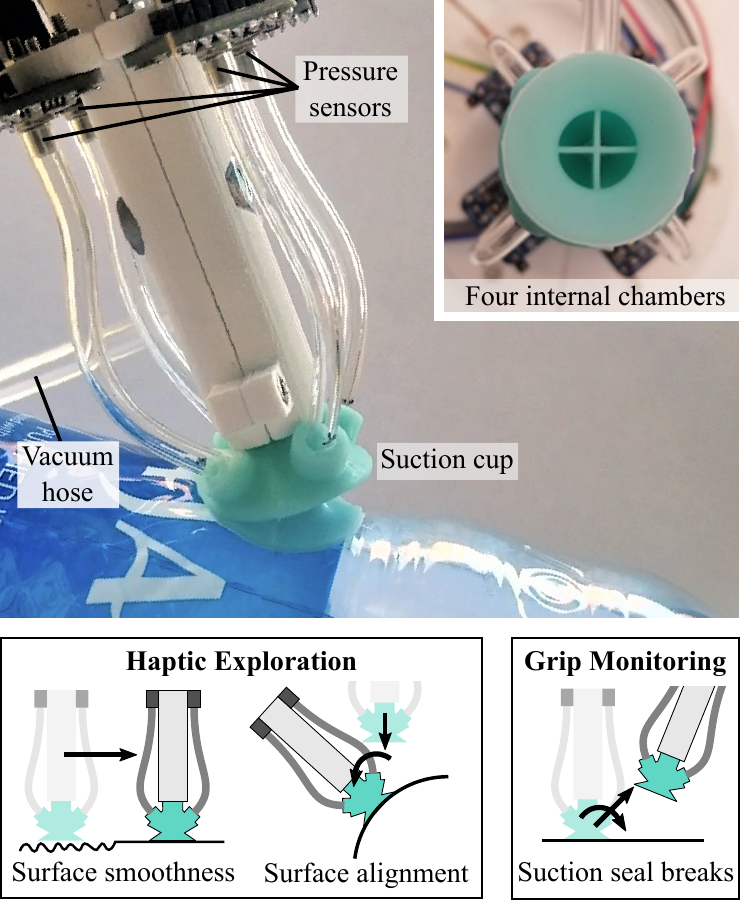}
% 	\includegraphics[width=1\linewidth]{./Figures/1_frontImage.png}
	%\end{subfigure}
    %\vspace{+1pt}
	\caption{The multi-chamber smart suction cup grips a water bottle. The cup has four internal chambers, each connected to a pressure transducer that provides a measure of internal flow rate. It is able to localize small breaks in the seal due to, for example, the rugosity (e.g., wrinkles, bumps, etc.) of the object surface or the application of external wrenches. Overall vacuum pressure is modulated in order to achieve different exploratory haptic procedures, such as sliding across surfaces. }
	\label{fig:main}
	\vspace{-15pt}
\end{figure}
%%%%%%

Prior tactile sensors designed for use in suction cups provide partial information about object properties and vacuum sealing state. Aoyagi et al. coat piezoresistive polymer on a bellows suction cup to measure compression forces \cite{aoyagi2020bellows}. Doi et al. implement a capacitive proximity sensor on the base plate of a suction cup end-effector to measure the distance from the plate to the object surface\cite{doi2020novel}. These methods measure vacuum state indirectly from the deformation of the suction cup and proximity to the object. Another straightforward approach is to monitor internal vacuum pressure of the suction cup as a discrete measure of suction sealing, as in \cite{eppner2016lessons}. None of these methods localize the source of a leak or measure local surface geometry. Nadeau et al. demonstrate how continuous suction-flow-rate measurement at the fingertips of a multifinger robotic hand can inform grasping and in-hand manipulation under water and note sensitivity to contact geometry \cite{nadeau2020tactile}, but do not address application to suction cups. 

\textit{We hypothesize that the continuous measurement of flow rate within the suction cup itself provides new opportunities for smart suction cup dexterity.}
% Another tactile sensor uses piezoresistive polymer 
% However, this approach does not provide directional local contact information of the suction cup.   
% For the uncertainty of object information, we can consider the tactile sensor, but in the literature, most of the sensors are not direct local sensing. In Amazon Challenge, pressure sensor could be a useful tool, but it only tells global vacuum success. Another example is using the piezo-resistive sensor → multiple suction cups, not a single suction cup manipulation. Capacitive proximity sensor also used to estimate the object relative distance in different direction, however, on both demonstrations, it could not measure the local seal information of the suction cup which should be critical for realtime control of the gripper.
In this paper, we present a novel sensing method using %suction cups that monitors local contact conditions, or the distribution of vacuum sealing around the outer perimeter of the cup. This suction cup contains four separated 
inner chambers (\cref{fig:main}), each of which connects to a pressure transducer to estimate distributed flow rates. %We now show how suction-flow-rate sensing is useful for manipulation tasks in air when applied with this multi-chamber suction cup design. %We use this directional suction airflow through each chamber to estimate the local contact states; a similar concept was demonstrated in an underwater suction gripper\cite{nadeau2020tactile}. 
In Section \ref{sec:DesignFab}, we describe the design and fabrication process of the suction cup, including Computational Fluid Dynamics (CFD) simulations of contact cases. Experimental results during haptic exploration in Section \ref{sec:Haptic} suggest that this sensor can detect transitions between surface textures by gentle sliding motions inspired by human haptic exploratory procedures \revision{(EP), e.g., prototypical finger rubbing} \cite{lederman1993extracting}. This design also enables the evaluation of geometric surface normal contact through palpation. In Section \ref{sec:Detach}, we monitor the suction signal during astrictive gripping as twists, resulting in external grasp wrenches, are applied. We find this sensor predicts the occurrence and location of catastrophic breaks in the suction seal around the perimeter of the cup. As discussed in Section \ref{sec:conclusion}, the strength of this smart suction cup is in the handling unknown objects with variable surface properties that may compromise sealing. % with a spatial resolution of 30deg with more than 97\% accuracy. % As discussed in Section \ref{sec:conclusion}, these results indicate that this sensing modality holds the potential to inform astrictive grasp planning and execution, especially for the suction-cup handling of unknown objects with hard-to-predict properties like texture, rugosity, porosity, and mass distribution.

%and dynamic detaching contact monitoring. It also identified spatial vacuum seal formation with  
%We also used the suction gripper to estimate the contact states of the vacuum seal when applied detaching wrenches. 
%Our tactile sensor should be able to empower a suction cup gripper to grasp and manipulate objects with various textures, geometry, and unknown inertia property (\cref{fig:main}).

\section{Design and Fabrication}
\label{sec:DesignFab}

\begin{figure}[tbp!]
	\centering
	\begin{subfigure}[h]{0.48\linewidth}
	\centering
	\includegraphics[width=\linewidth]{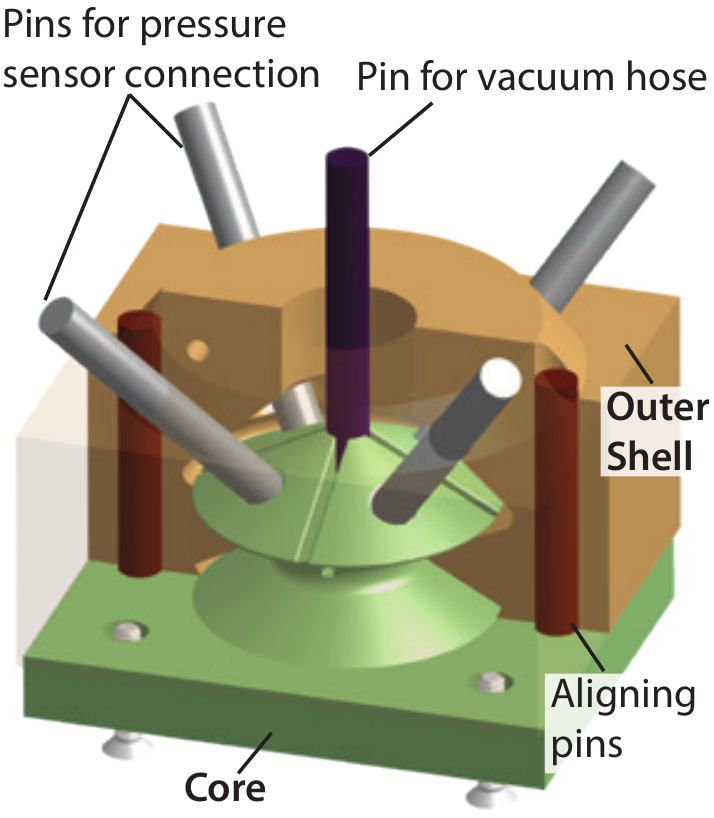}
	\caption{Casting mold}
	\label{fig:mold}

    \end{subfigure}
\hfill
    \centering
	\begin{subfigure}[h]{0.50\linewidth}
	\vspace{15pt}
	\includegraphics[width=\linewidth]{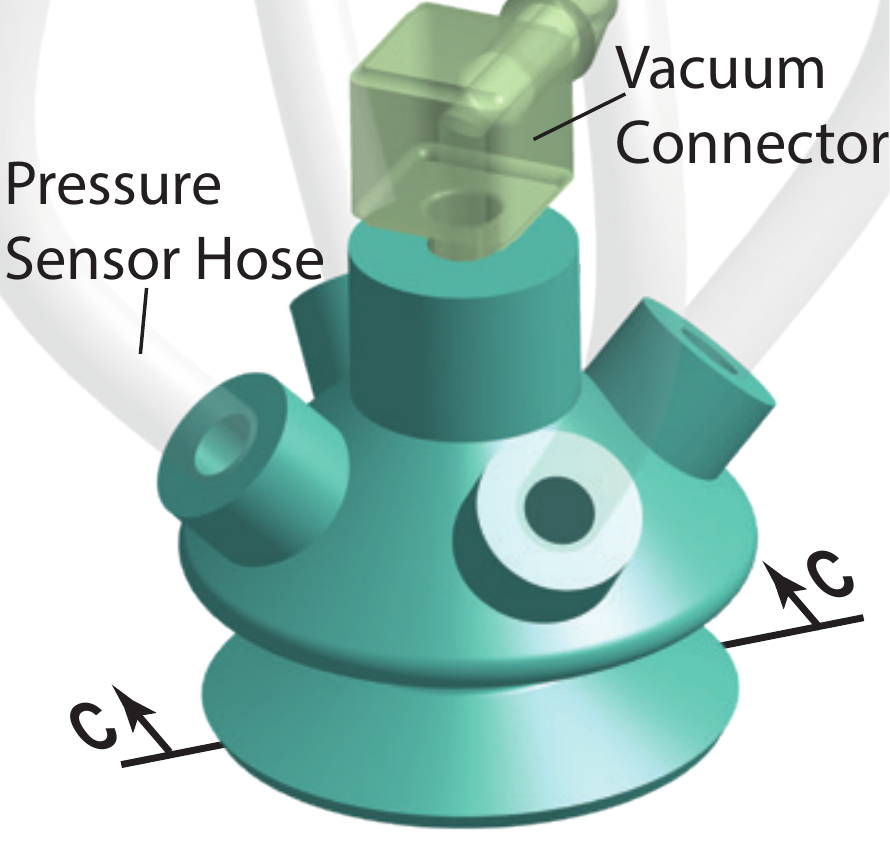}
	
	\caption{Suction cup model}
	\label{fig:castingResult}
    \end{subfigure}
  
    \begin{subfigure}[h]{1\linewidth}
    \centering
	\includegraphics[width=0.75\linewidth]{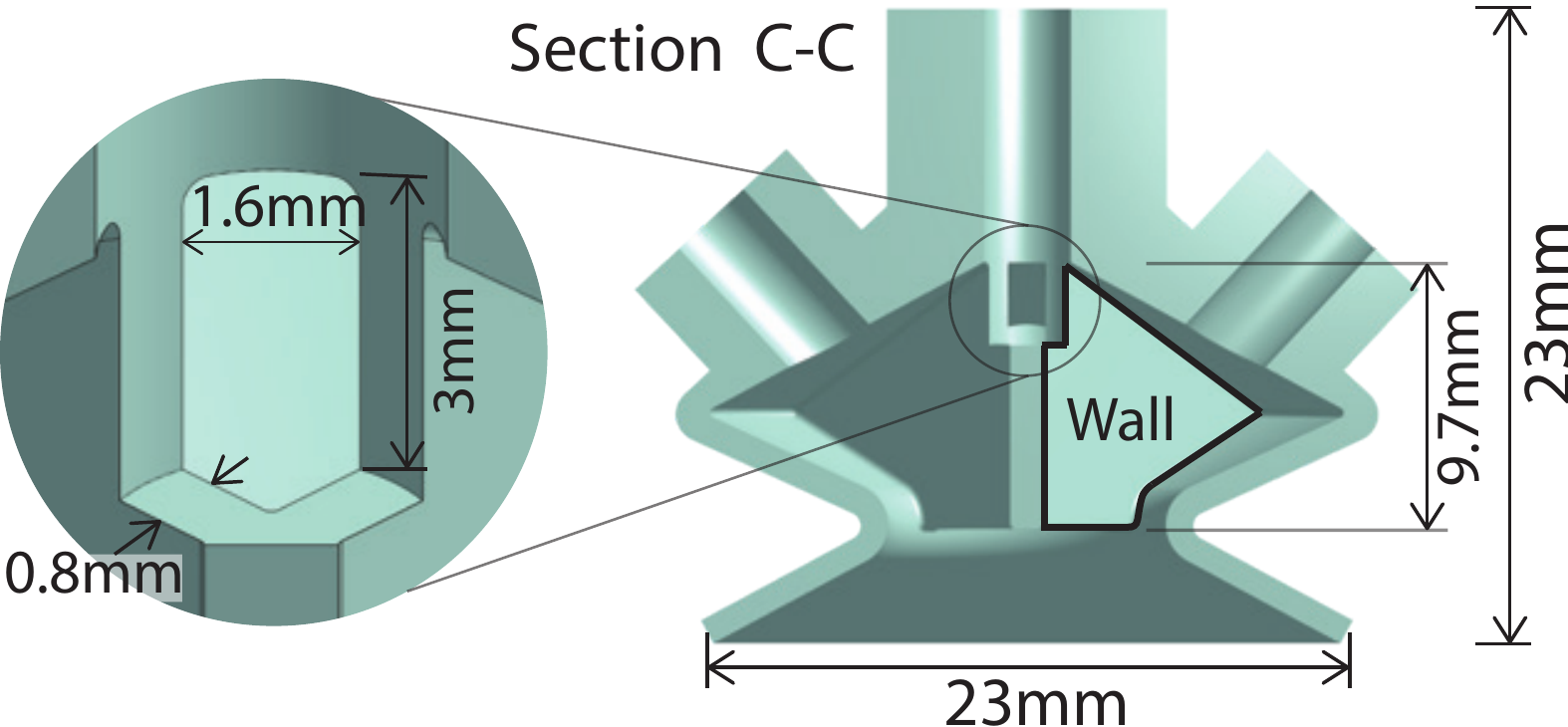}
% 	\vspace{1pt}
	\caption{Suction cup cross-section view}
	\label{fig:cut3D}
    \end{subfigure}
%\hfill 
%\null
	\caption{Casting mold and design of the suction cup. (a) Casting mold has three parts (2 Outer shells and 1 core). Molds are aligned and fixed by pins and bottom bolts. (b) The resulting suction cup is connected with vacuum connector and hoses to the pressure sensors. (c) Cross-sectional view of the suction cup shows internal and outer key dimensions.}
	\label{fig:Fabrication}
	
	\vspace{-15pt}

\end{figure}

This smart suction cup utilizes airflows inside its chambers to monitor local contacts. %Hannah: let's consolidate this citation in the introduction. %Suction water flow rate could estimate contact states (e.g., the contact angle of a fingertip) of an underwater suction gripper \cite{nadeau2020tactile}, and we extend this concept to the air suction flow. 
Internal wall structures separate the suction cup into four chambers (\cref{fig:main}). Suction airflow is separated into each chamber and the pressure sensor connected to each chamber provides an estimate local flow rate. \revision{Studying how spatial resolution and gripping performance is affected by increasing the number of chambers will be a future work.} We implement the wall structure inside a single-bellows suction cup (\cref{fig:Fabrication}) for its versatility on different curvatures and orientations of objects. The internal wall structure only spans the proximal portion of the suction cup, in order to maintain typical flexibility, deformation and seal formation in the distal lip. 

We perform fabrication, including the chamber walls, with a single-step casting of silicone rubber. The casting mold is comprised of three parts, two outer shells and one core (\cref{fig:Fabrication}), that are 3D printed using an SLA 3D printer (Formlabs, Form2). To ensure the clean casting of the thin wall structures (0.8mm thick), we used a syringe with a blunt needle (gauge 14) to inject uncured RTV silicone rubber (Smooth-On, MoldMax 40) and vacuum-degassed it. After curing, the outer shells are removed and the silicone suction cup is stretched and peeled off of the inner core mold. Tearing of the silicone can occur during this step, especially with harder rubbers.

% \begin{figure}[tbp!]
% 	\centering
% 	\begin{subfigure}[h]{0.47\linewidth}
% 	\centering
% 	\includegraphics[width=\linewidth]{./Figures/2_1_moldCad.pdf}
% 	\caption{Casting Mold}
% 	\label{fig:mold}
	
% 	\vspace{5pt}

%     \end{subfigure}
% \hfill
%     \centering
%     % 
% 	\begin{subfigure}[h]{0.51\linewidth}
% 	\vspace{26pt}
% 	\includegraphics[width=\linewidth]{./Figures/2_2_moldResult_ver2.pdf}
% % 	\vspace{1pt}
% 	\caption{Suction cup cross-section view}
% 	\label{fig:cut3D}

% % 	\centering
% % 	\includegraphics[width=\linewidth]{./Figures/2_3_onFixture.pdf}
% % 	\caption{Suction cup attached to the 3D printed fixture}
% % 	\label{fig:onFixture}
%         \end{subfigure}
% %\hfill 
% %\null
% 	\caption{Fabrication of the suction gripper. \han{Hannah: can we have a more descriptive caption? Label the terms that you have in the main text: outer shell and core. Do the alignment pins go into the holes of the green part? If so, it's odd that they don't align in image. If space isn't an issue, then a fabrication sequence would be a great addition -- perhaps this is added to TRO version. Also, I'd like to have a label on 2(b) showing the internal wall in the proximal portion of the suction cup. Perhaps outlining one of the walls and then label? Can also highlight where the the vacuum inlet of the chamber is (small rectangle at top). You may also be able to show the cross-section plane on 2(a) or a small inset of the actual cup in (b)... it's not easy to discern right now.}}
% 	\label{fig:Fabrication}
	
% 	\vspace{-15pt}

% \end{figure}

\subsection{System integration}
\label{sec:system}

%\han{A little confusing, so I tried rewording based on my understanding. Please correct new mistakes. }

%We integrated the fabricated suction cup with a vacuum generator, pressure sensors, and a robot arm. 
The suction cup is connected with pressure transducers and a vacuum source with pressure regulation. For experimental trials, this system is then integrated with a Universal Robot arm. We used ROS to control the robot arm and collect both pressure sensor and wrist force/torque (F/T) sensor data.

Four ported pressure sensors (Adafruit, MPRLS Breakout, 24bit ADC, 0.01Pa/count with an RMS noise of 5.0 Pa) connect with the four chambers of the smart suction cup via polyurethane tubes. %For the pressure sensor connected to each internal chamber, we used a ported pressure sensor (Adafruit, MPRLS Breakout, 24bit ADC, 0.01Pa/count) that has an RMS noise of 5.0 Pa. 
%Pressure sensors are remotely connected to the suction cup via a polyurethane tube, allowing it to remain small and flexible. 
The suction cup and the pressure sensors attach to a 3D printed fixture (\cref{fig:main}). This fixture is then attached to the wrist F/T sensor (ATI, Axia80, sampling rate 150Hz) on the robot arm (Universal Robots, UR-10). A microcontroller (Cypress, PSoC 4000s) is fixed to the arm proximal to the load cell and communicates with the four pressure sensors via I2C at a 166.7Hz sampling rate. 

A vacuum generator (VacMotion, VM5-NA) converts compressed building air to a vacuum source with a maximum vacuum of 85kPa. A solenoid valve (SMC pneumatics, VQ110, On/off time = 3.5 / 2ms), commanded by a microcontroller, regulates the compressed air as a means of moderating vacuum intensity. During haptic exploration experiments, the valve is controlled with pulse width modulation (PWM) at a frequency of 30\,Hz with 30\% duty cycle to approximate lower vacuum pressures. We chose this PWM setting considering the on-off time of the solenoid valve and the sampling rate of the pressure sensor. % The generator is connected to a regulated compressed air (70kPa) through a solenoid valve (SMC pneumatics, VQ110, On/off time = 3.5 / 2ms). we used this fast switching valve to implement PWM control to the vacuum pressure for the haptic exploration experiments. 
The vacuum hose that applies suction to the cup is attached at both the suction cup vacuum connector and proximal to the load cell to reduce tube movement and subsequent F/T coupling.

%The microcontroller and the vacuum hose are attached to the base of the wrist joint to minimize the F/T coupling from contacts other than the suction cup. 
%Another microcontroller of the same kind was used to control the solenoid valve. 

\subsection{CFD Simulation}
\label{sec:DesignCFD}

\begin{figure}[tbp!]
	\centering
	\begin{subfigure}[h]{0.45\linewidth}
	\centering
	\includegraphics[width=\linewidth]{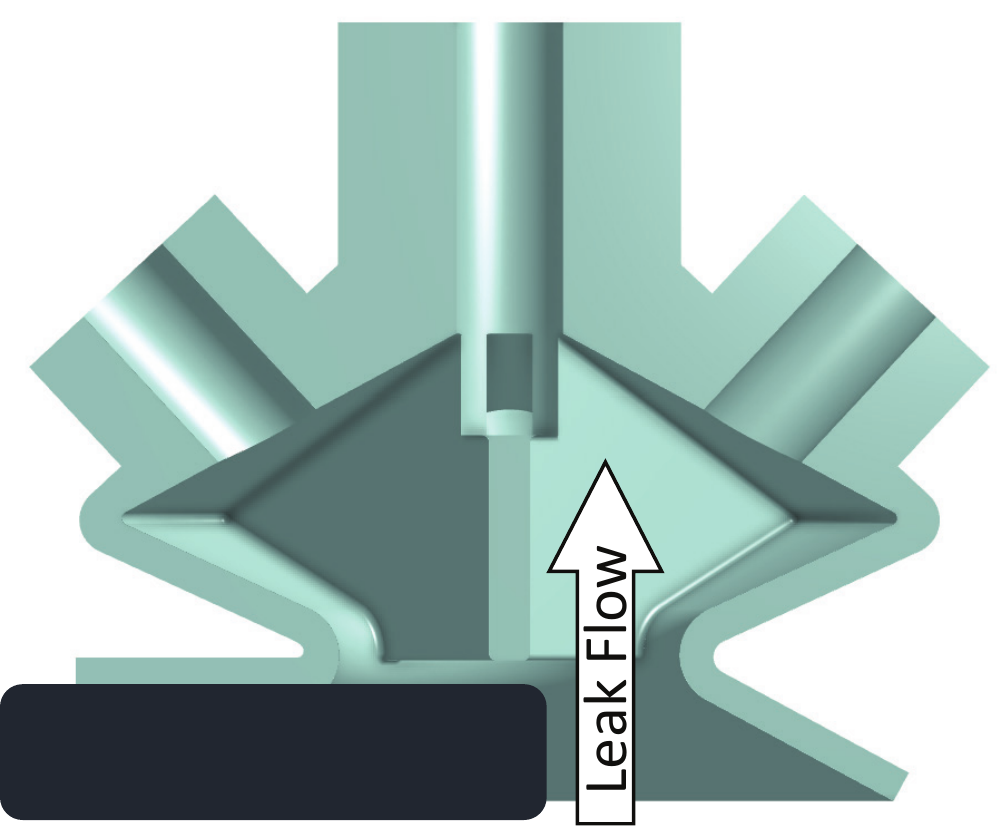}
	\caption{Vertical leakage airflow}
	\label{fig:Schem_vertical}
        \end{subfigure}
%\hfill
	\begin{subfigure}[h]{0.45\linewidth}
	\centering
	\includegraphics[width=\linewidth]{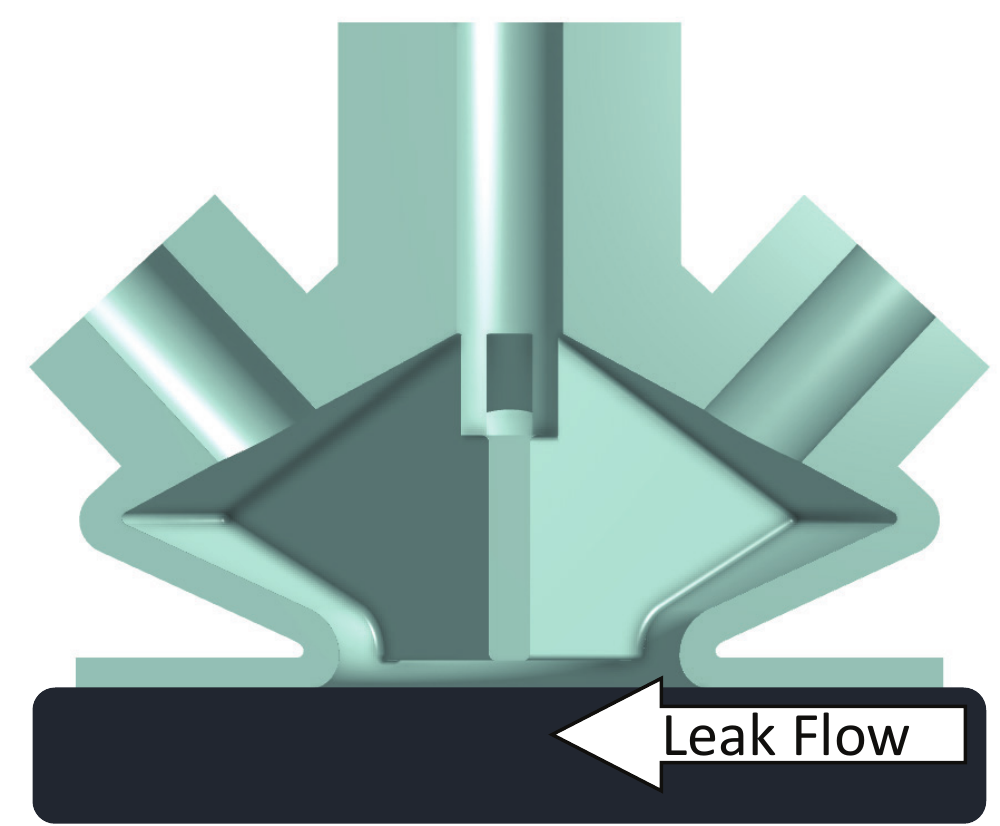}
	\caption{Horizontal leakage airflow}
	\label{fig:Schem_horizon}
        \end{subfigure}
%\hfill 
%\null
	\caption{Two cases of CFD simulation. Dark blocks are engaged objects.}
	\label{fig:ComsolAssumption}
\vspace{-10pt}
\end{figure}

\begin{figure}[tbp!]
	\centering
	\begin{subfigure}[h]{0.49\linewidth}
	\centering
	\includegraphics[width=\linewidth]{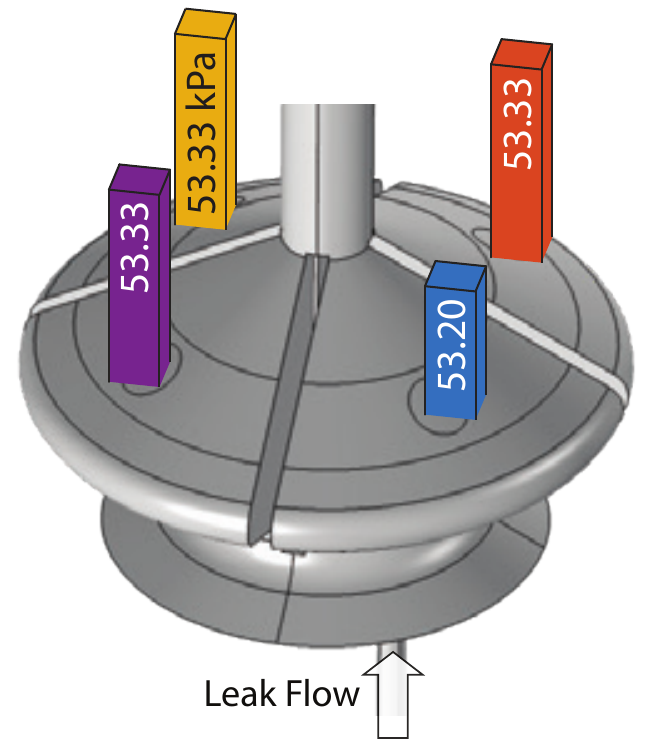}
	\caption{Vertical Orifice}
	\label{fig:VerticalOrif}
        \end{subfigure}
%\hfill
	\begin{subfigure}[h]{0.49\linewidth}
	\centering
	\includegraphics[width=\linewidth]{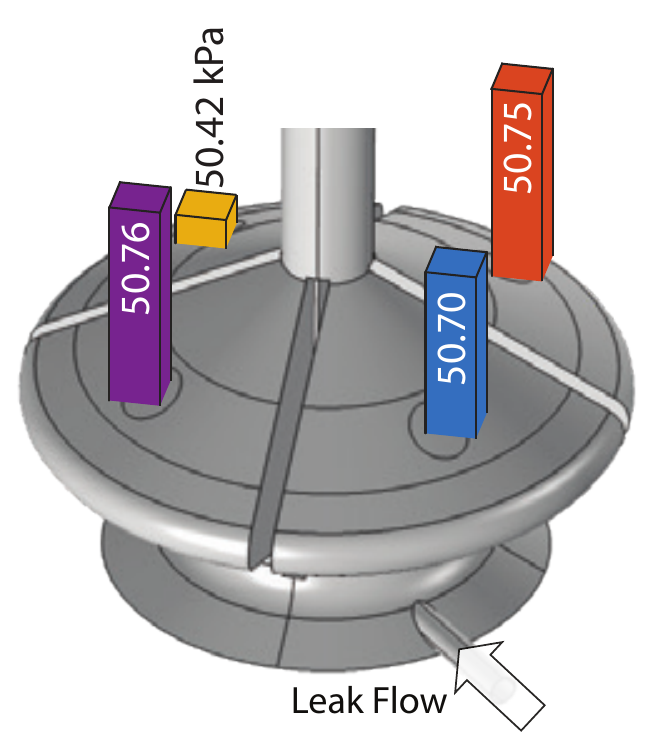}
	\caption{Horizontal Orifice}
	\label{fig:horizontalOrif}
        \end{subfigure}
%\hfill 
%\null
	\caption{CFD result of the $P_{vac}$ measured at the sensor locations of each chamber. Bar graphs are scaled to represent the 0.4kPa from the maximum of the four $P_{vac}$s}
	\label{fig:ComsolOrifs}
\vspace{-10pt}
\end{figure}

\begin{figure}[tbp!]
	\centering
	\begin{subfigure}[h]{0.50\linewidth}
	\centering
	\includegraphics[width=\linewidth]{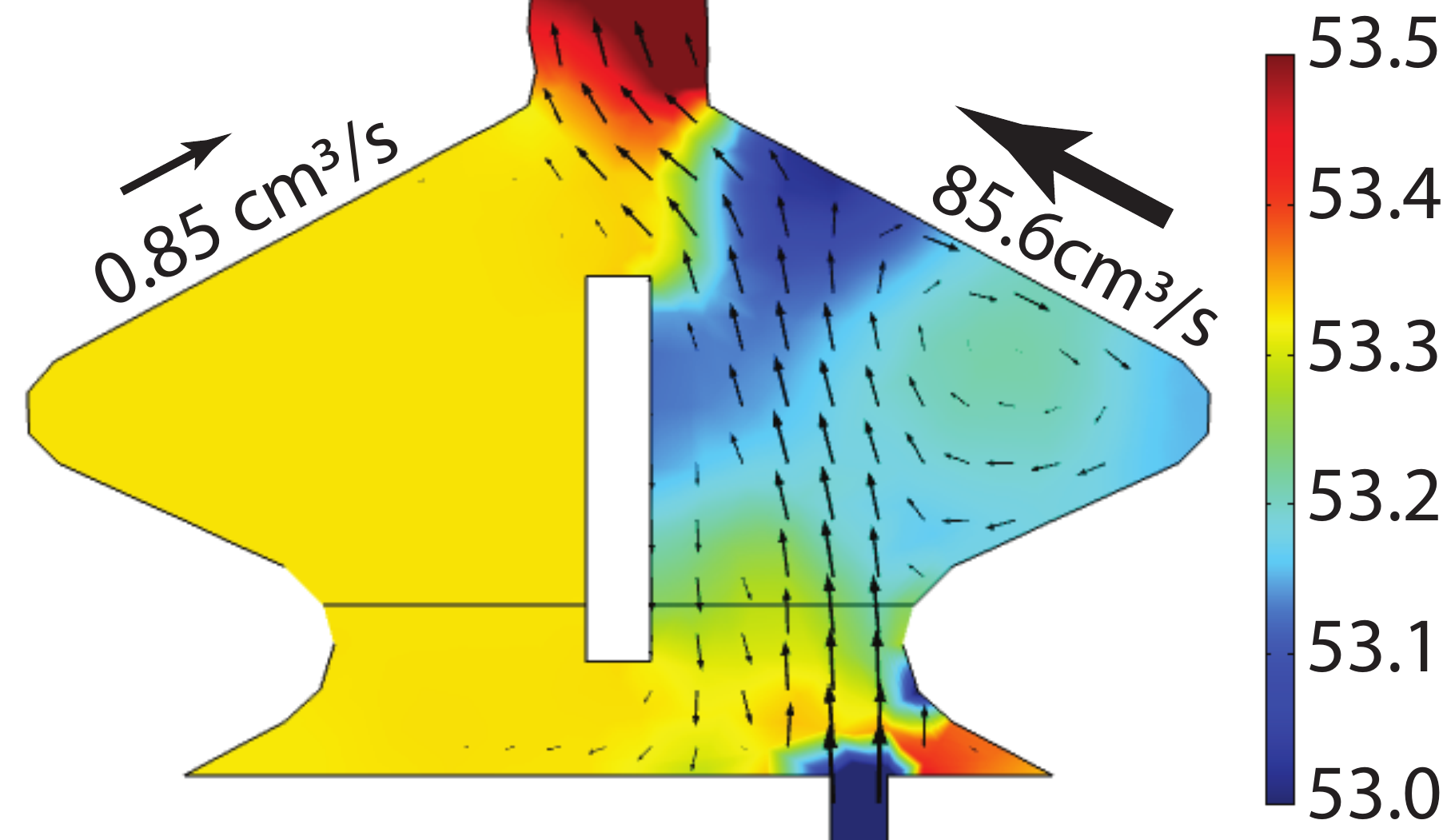}
	\caption{Vertical Orifice}
	\label{fig:Vertical2D}
        \end{subfigure}
%\hfill
	\begin{subfigure}[h]{0.48\linewidth}
	\centering
	\includegraphics[width=\linewidth]{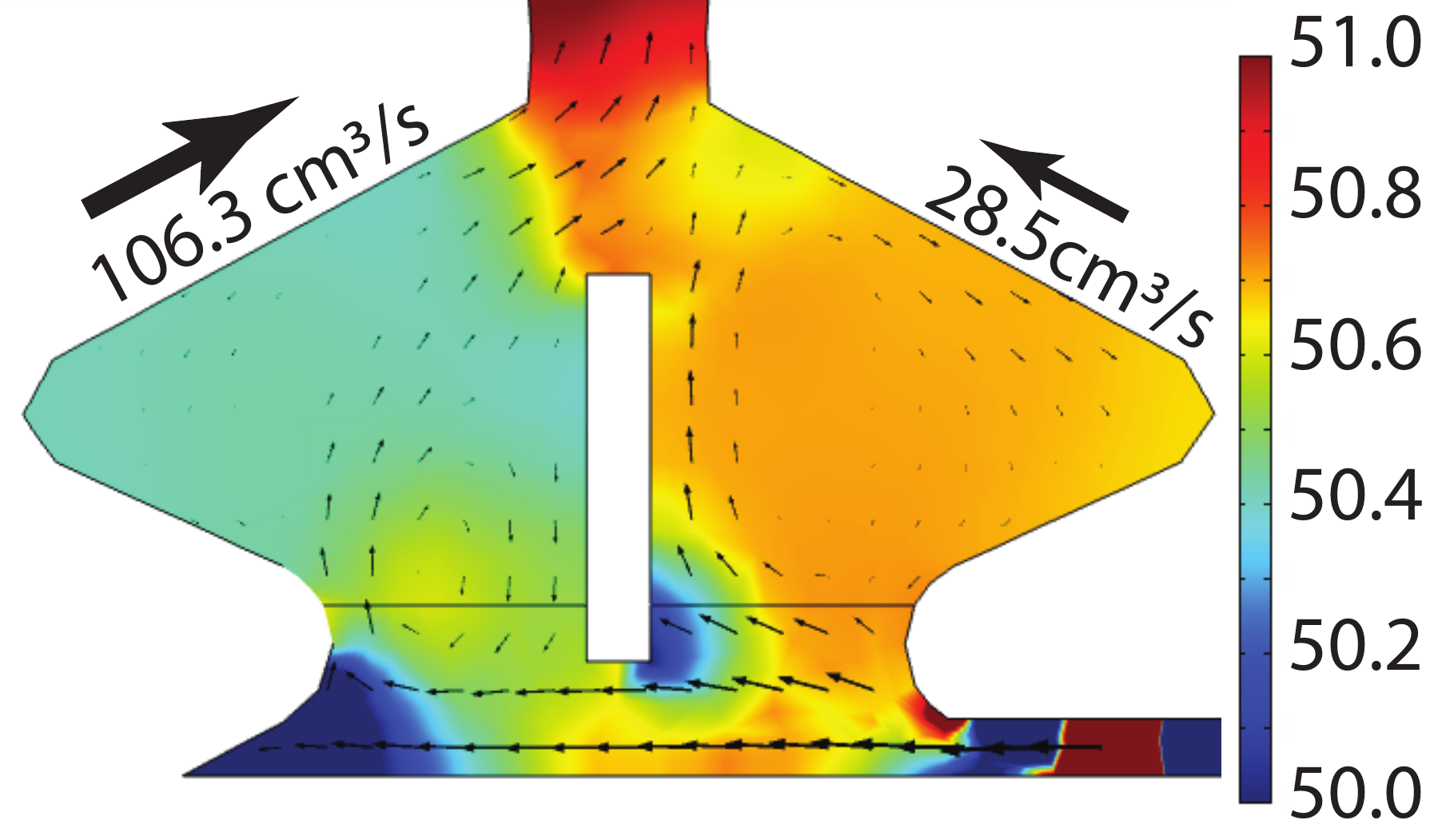}
	\caption{Horizontal Orifice}
	\label{fig:horizon2D}
        \end{subfigure}
%\hfill 
%\null
	\caption{Cross sectional view of the pressure distribution. Arrows inside represents relative logarithmic scale of air flow velocity. The colormap unit is kPa.}
	\label{fig:comsolHeatMap}
\vspace{-10pt}
\end{figure}

Using Computational Fluid Dynamics (CFD) simulation (COMSOL Multiphysics, $k-\epsilon$ turbulence model), we evaluate the gripper in two \revision{example} suction flow cases: vertical and horizontal flow (\cref{fig:ComsolAssumption}). The vertical flow case emulates when the suction cup only partially contacts a surface, or when the surface's shape inhibits sealing. %In haptic explorations, especially during geometric search, we surmise that the contact will often be partially made, flowing the leakage air vertically. 
However, when the suction cup engages with a smooth flat surface, flow can only move inward from the outer edges of the cup, as in the horizontal flow case. This horizontal leak is common as the suction cup is wrenched from the surface after a suction seal is formed. %, the vacuum seal is assumed to break from a single direction, flowing the leakage air horizontally along the contact surface. 
Although the suction cup will deform under vacuum pressure, we use modeled rigid geometry in the CFD simulation. For each case, we %used the same geometry as our designed suction cup and 
approximate the leak flow direction with a small pipe (D = 1mm, L = 7mm) located close to one of the internal chambers as shown in \cref{fig:ComsolOrifs} and \cref{fig:comsolHeatMap}. The boundary conditions of the vacuum pump pressures and flow rates match the experimental setup.

The simulation results suggest that the gripper can locate leakage flow using differences between the four pressure transducers. We defined vacuum pressure ($P_{vac}$) as 
\begin{equation}
% \vspace{-15pt}
P_{vac} = P_{atm} - P_{chamber} .
% \vspace{+5pt}
\label{eqn:P_vac_Def}
\end{equation}
In the vertical leakage flow case, $P_{vac}$ close to the leaking orifice shows the least vacuum pressure than the others (\cref{fig:VerticalOrif}). On the other hand, the horizontal leakage causes the diagonally opposite channel %across the center 
to have the lowest $P_{vac}$ (\cref{fig:horizontalOrif}). %These two different pressure distributions are caused by the flow rates through each chamber. 
These trends are supported by the flow results in \cref{fig:comsolHeatMap}, where the vertical and horizontal orifices produce the highest flow rate is opposite chambers.
%In the vertical leakage case, as shown in \cref{fig:Vertical2D}, the flow mostly passes through the orifice chamber while in the horizontal case (\cref{fig:horizon2D}), more leakage flow passes through the chamber across due to the flow path along the bottom surface. 
The simulation result also shows an estimate of the pressure difference between chambers ($\sim$0.4kPa) which can be differentiated by the selected pressure sensors.
\revision{Despite the two distinctive results of these idealized simulations, %the actual pressure distributions may not be interpretable due to the 
real suction cup deformation and complex leak geometry necessitate data-driven analysis of pressure as a tactile sensor as described in \cref{sec:Detach}.}

% F/T sensor + data to compensate the orifice angle dependency.show in section XX

% We To provide intuition of how the airflow and pressure distribution will be in different types of leakage flow. In practice, the airflow will be more complex because of the the deformation and the randomly formed break point. That is why we rely on the data driven method to estimate the break point rather than modeling every possible leakages. 
\section{Haptic Exploration}
\label{sec:Haptic}

In suction cup grasping, contact surface smoothness enables the formation of %is critical in forming 
a suction seal. Surface textures and rugosity alter the quality of the seal that determines the resistible grasp wrench. Local geometry of the contact, which is influenced by surface curvature and suction cup orientation, can also compromise seal formation. %, because the suction cup may not be able to form a vacuum seal if the geometry is too sharp or the suction cup is misaligned to the surface normal. 
However, this information is difficult to obtain from vision alone because of limited resolution and potential for occlusion. %Therefore haptic exploration at the direct contact should be an effective solution to examine critical object properties. 
% Force/Torque sensors on the wrist can also provide useful information about local geometry; however, in a cluttered environment, the end-effector may contact multiple adjacent objects, making the contact geometry estimation erroneous.

We propose a method of haptic exploration using lower suction forces by regulating vacuum pressure. Using full vacuum power %for the exploration could be possible, however, it can 
can apply large suction forces, even to non-optimal contacts, causing the object to move and making it difficult to move the cup to a new location. %causing unwanted disturbance of the object pose. 
Using lowered vacuum pressure, the suction cup can gently slide along the surface, allowing efficient continuous exploration. %Therefore, we demonstrate haptic exploration methods by lowing the vacuum pressure using PWM control of the solenoid valve.
%%%%%

\subsection{Experiments}

First, we estimate surface texture using the smart suction cup. On a substrate with sandpapers of different roughness (120 to 600 grit), we apply both full vacuum and lowered vacuum to the suction gripper and measure the pressure sensor response. %In order to achieve the lower vacuum, we apply 30Hz PWM with 30\% duty cycle. We chose this PWM setting considering the on-off time of the solenoid valve and the sampling rate of the pressure sensor.  Hannah: Moved this detail to Implementation so we only report it once.
For both full vacuum and lower vacuum cases, the robot arm approaches the contact surface without a vacuum until the normal force reading from wrist F/T sensor reaches 0.5N, and then the vacuum is turned on.

\begin{figure}[tbp!]
    \centering  
	\begin{subfigure}[h]{0.49\textwidth}
    	\centering
    	\includegraphics[width=0.8\linewidth]{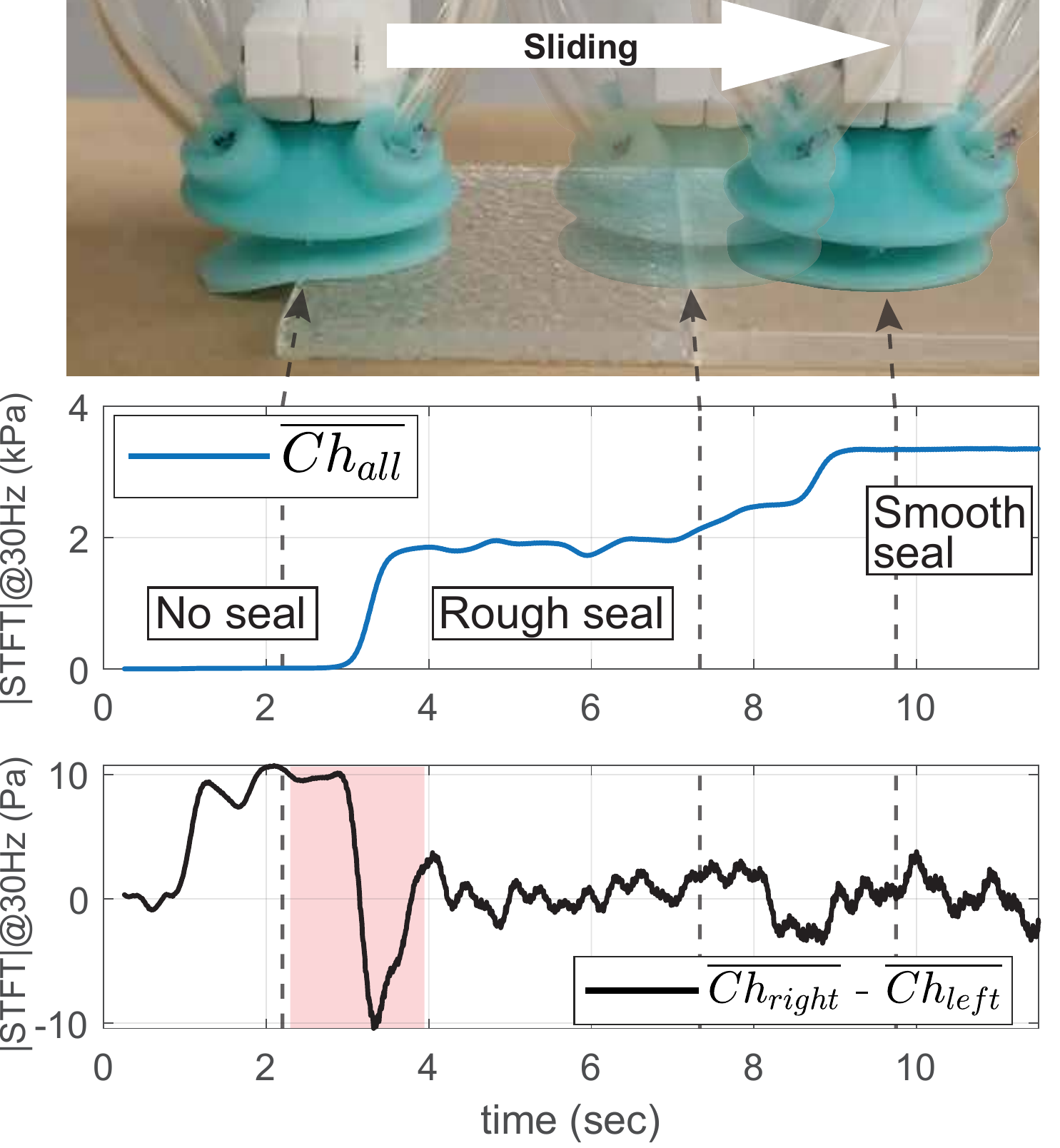}
    	\caption{Wavy-smooth acrylic}
    	\label{fig:sliding_Wavy}
    	\vspace{+8pt}
	\end{subfigure}
    \centering
	%\vspace{-10pt}
	\begin{subfigure}[h]{0.49\textwidth}
	    \centering
	    \includegraphics[width=0.8\linewidth]{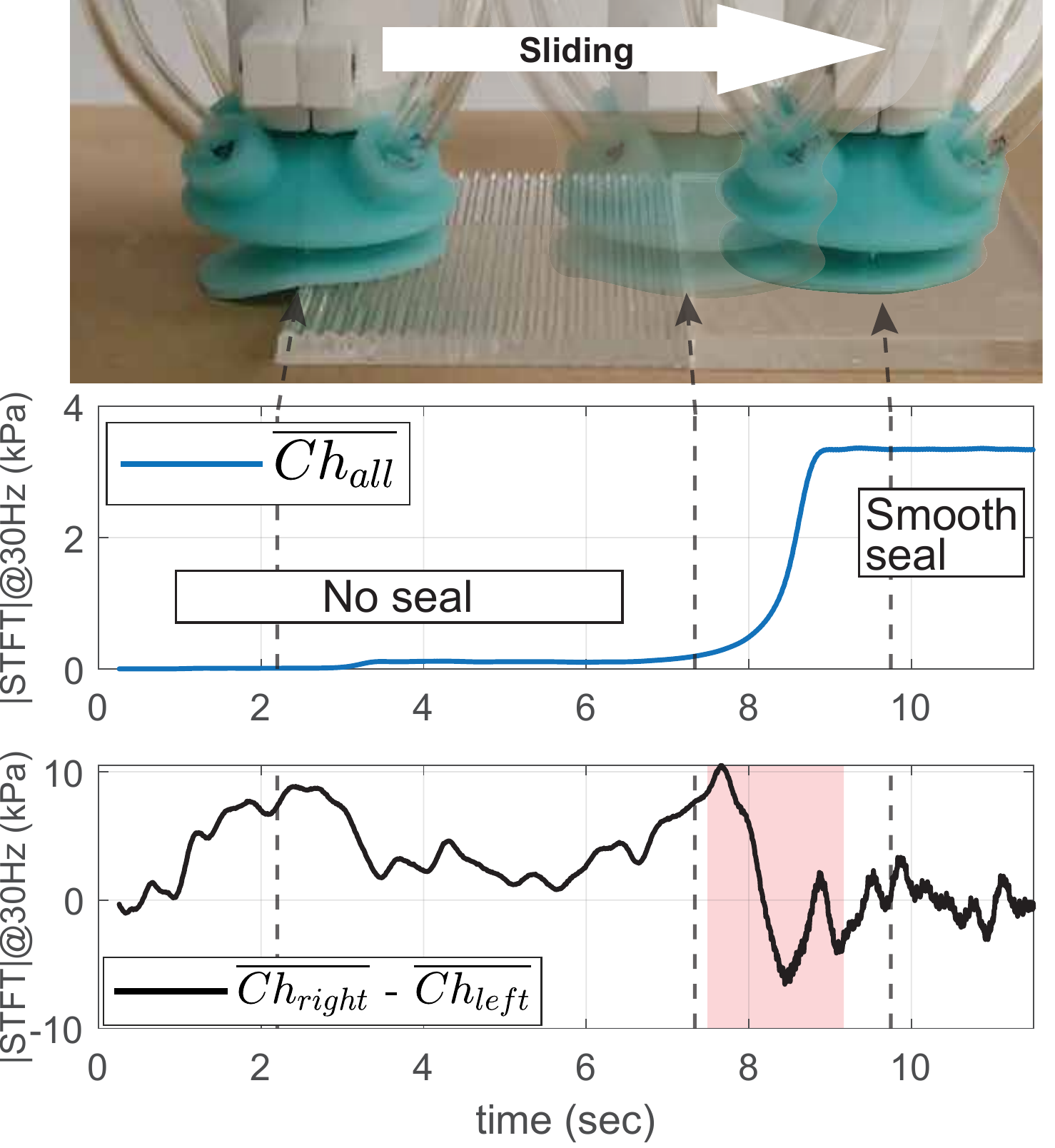}
		\caption{Ribbed-smooth acrylic}
		\label{fig:sliding_Ribbed}
		\vspace{+5pt}
	\end{subfigure}
    
	\caption{Sliding test results on (a) wavy-smooth acrylic and (b) ribbed-smooth acrylic pairs. (Top) Snapshot of the sliding motion at the time marked with dashed line. (middle) Average $|STFT_{30}|$ of all four channels (bottom) Difference of $|STFT_{30}|$ between right two channels and left two channels. Red-shades are when large transition (no seal to seal) occurs.}
		
	\vspace{-15pt}
	\label{fig:slidingAll}
\end{figure}

Second, we test sliding haptic explorations on acrylic plates with textured-smooth surface transitions (\cref{fig:slidingAll}). Using the robot arm, we apply an initial normal force of 0.5N and then apply low suction. We exclude the full vacuum setting because it inhibits sliding. %drive the solenoid valve with PWM control of 30Hz, 30\% duty cycle. 
The suction cup starts with only half contact on the textured surface. The robot then moves the suction cup laterally at 6.5 mm/sec speed while measuring the vacuum pressure in each chamber. We test on two sets of surface transitions: wavy-smooth and ribbed-smooth. We laser-cut acrylic plates in 35mm x 51mm and bonded textured-smooth acrylic pairs along the longer edge using acrylic cement.

% (TAP plastics, Innsbruck(wavy) and Ribbed)
% We used acrylic instead of sand paper to avoid abrasion.
% and monitors the sensor signal change as it transitions from textured acrlyic to smooth acrylic.
% to search better surface to engage

\begin{figure}[tbp!]
\centering
	%\vspace{-10pt}
	%\begin{subfigure}[h]{0.5\textwidth}
	%\centering
	\includegraphics[width=0.8\linewidth]{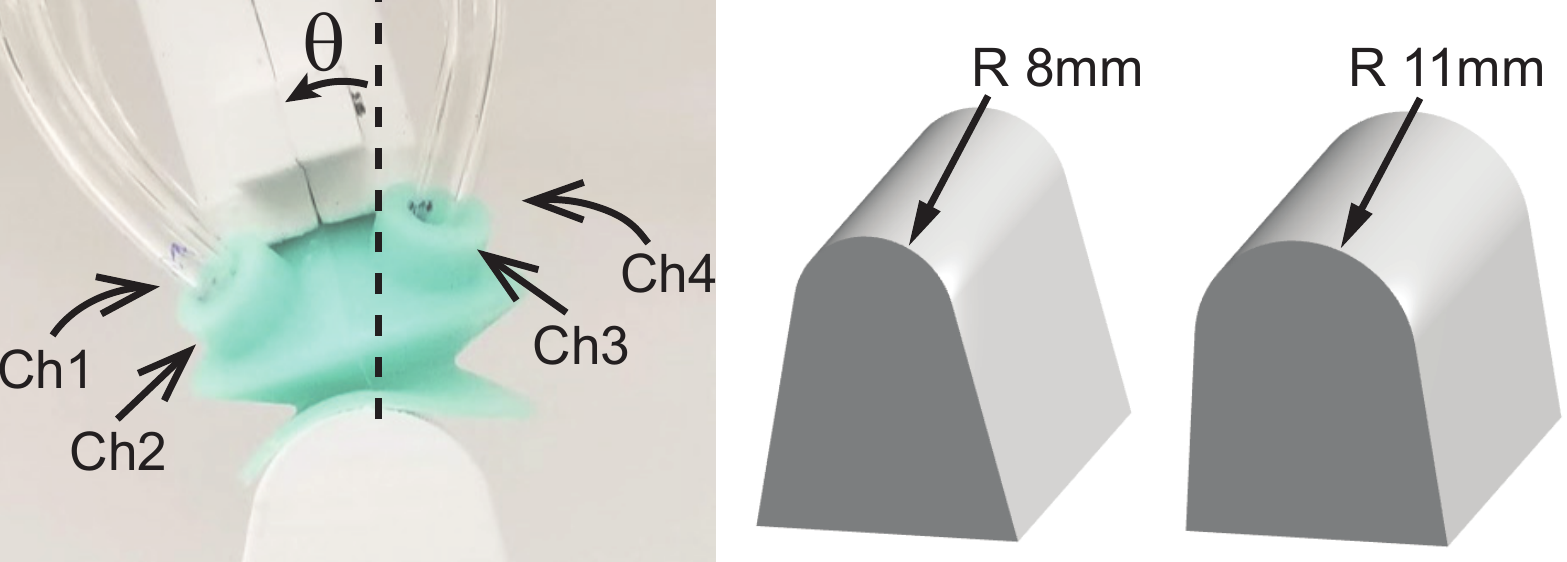}
	%\end{subfigure}
    \vspace{+1pt}
	\caption{Experiment setup for seeking a surface normal on two different objects}
	\label{fig:surfaceNormalSetting}
	\vspace{-15pt}
\end{figure}

Third, we use our tactile sensing suction gripper to seek surface normal and evaluate surface curvature through palpation. We use two 3D printed test objects with different contact radii at the tip (8mm and 11mm), shown in \cref{fig:surfaceNormalSetting}. %Around the contact point of the test object, t
The robot changes the orientation of the end-effector from normal by pivoting about the tip of the object with a range of $\pm 30$deg in 15deg increments. At each orientation, the robot then translates the suction cup until the gripper normal force reaches a set threshold, either 0.5N or 1N. First, we record preasure transducer readings at low vacuum, then we evaluate suction grasp at the surface normal with maximum vacuum. % and the pressure on with the same PWM control. 
%assuming the force controlled search, 
\begin{figure}[tbp!]
\centering
	%\vspace{-10pt}
	%\begin{subfigure}[h]{0.5\textwidth}
	%\centering
	\includegraphics[width=1\linewidth]{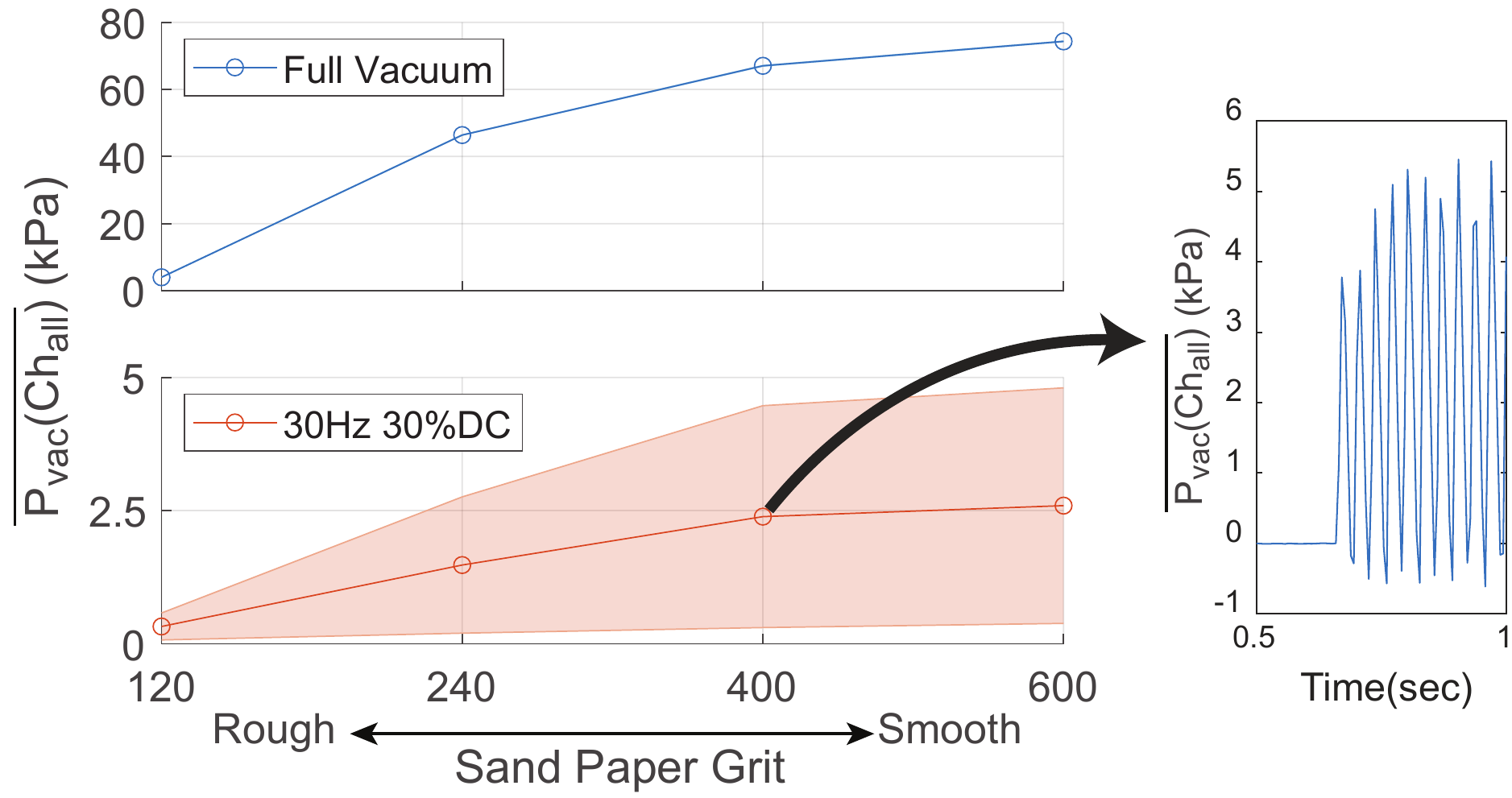}
	%\end{subfigure}
    %\vspace{+1pt}
	\caption{Texture evaluation result with full vacuum (top) and lowered vacuum (bottom) mode. For lowered vacuum, the valve was controlled with 30Hz, 30\% duty cycle PWM. Shades represent $\pm1$ standard deviation. (Right) $P_{vac}$ on 400 grit sand paper with lowered vacuum.}
	\label{fig:TwoModes}
	\label{fig:One example}
	%\vspace{-5pt}
%\end{figure}
%%%%%%
%%%%%
% \begin{figure}[tbp!]
% \centering
% 	%\vspace{-10pt}
% 	%\begin{subfigure}[h]{0.5\textwidth}
% 	%\centering
% 	\includegraphics[width=0.9\linewidth]{./Figures/6_oneExample.pdf}
% 	%\end{subfigure}
%     %\vspace{+1pt}
% 	\caption{Vacuum pressure measured on 400 grit sand paper with PWM controlled vacuum. The pressure oscillates at PWM frequency (30Hz).}
% 	\vspace{-5pt}
% 	\label{fig:One example}

% \end{figure}
%%%%%%

%%%%%
%\begin{figure}[tbp!]
\centering
	%\vspace{-10pt}
	%\begin{subfigure}[h]{0.5\textwidth}
	%\centering
	\includegraphics[width=0.75\linewidth]{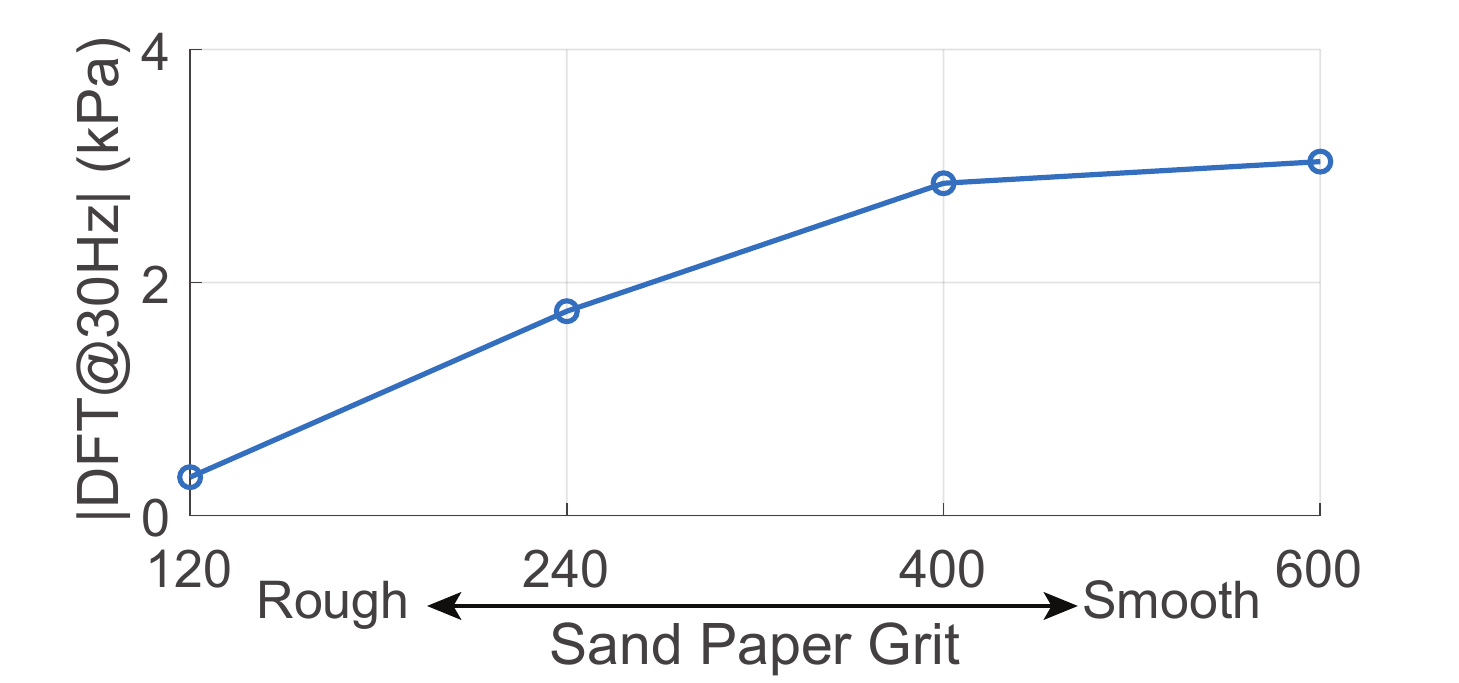}
	%\end{subfigure}
    %\vspace{+1pt}
	\caption{Magnitude of discrete Fourier transform of the lower vacuum mode at 30 Hz}
	\label{fig:DFT_surface}
	\vspace{-10pt}
\end{figure}
%%%%%% 

%%%%%%

\subsection{Results}
\subsubsection{\textbf{Surface Texture evaluation}}
While both full vacuum mode and lowered vacuum mode differentiate surface textures, the lowered vacuum mode interacts with the surface more gently. When engaged with the full vacuum mode, the measured $P_{vac}$ in the suction cup increases with smoothness (\cref{fig:TwoModes}(top)), reaching 74 kPa for the smoothest 600 grit sandpaper. %However, these high $P_{vac}$s will try to lift the object even on somewhat rough surfaces, which may disturb the current pose of the object and do not allow gentle haptic exploration. 
In contrast, the lowered vacuum mode produces 25 times lower $P_{vac}$ on average while still showing a similar trend between mean pressure and surface smoothness (\cref{fig:TwoModes}(bottom)). 

The high variance in $P_{vac}$ with the lower pressure mode is induced by the frequent on-off cycles in PWM control, resulting in poor signal-to-noise ratio.
%%%%%
% \begin{figure}[tbp!]
% 	\centering
% 	\begin{subfigure}[h]{0.50\linewidth}
% 	\centering
% 	\includegraphics[width=\linewidth]{./Figures/5_Comsol_Vertical.pdf}
% 	\caption{Vertical Orifice}
% 	\label{fig:Vertical2D}
%         \end{subfigure}
% %\hfill
% 	\begin{subfigure}[h]{0.48\linewidth}
% 	\centering
% 	\includegraphics[width=\linewidth]{./Figures/5_Comsol_Horizontal.pdf}
% 	\caption{Horizontal Orifice}
% 	\label{fig:horizon2D}
%         \end{subfigure}
% %\hfill 
% %\null
% 	\caption{Cross sectional view of the pressure distribution. Arrows inside represents relative logarithmic scale of air flow velocity. The colormap unit is kPa.}
% 	\label{fig:comsolHeatMap}
% \vspace{-10pt}
% \end{figure}
However, we find that the amplitude of this oscillating $P_{vac}$ signal provides a clean measure of surface texture. On smoother surfaces, the formation of a good suction seal allows higher variation in $P_{vac}$ between on and off phase, causing a higher amplitude of oscillation at PWM frequency (\cref{fig:One example} (right)). To effectively capture the amplitude of the signal, we use the magnitude of Discrete Fourier Transform (DFT) at the PWM driving frequency (30Hz)(\cref{fig:DFT_surface}), which shows similar trends as raw $P_{vac}$ measures in \cref{fig:TwoModes}(top). In the following subsections, we use DFT magnitude as a metric of contact evaluation.

\subsubsection{\textbf{Sliding exploration}}

% One adavantage of sliding is a continuous monitoring --> so we use STFT
To monitor gradual changes of the DFT during sliding motions, we use Short Time Fourier Transform (STFT) that uses Hamming windowed data of 0.5 seconds (83 samples). Each time stamp in \cref{fig:slidingAll} is evaluated as a center of this 0.5-second window. We define the magnitude of STFT at PWM driving frequency (30Hz) as $|STFT_{30}|$. Using this $|STFT_{30}|$ of the pressure sensors, the transition of the contact surface properties can be detected by sliding motions. In this test, contact transitions are compared with both the average $|STFT_{30}|$ over all chambers ($\overline{Ch_{all}}$) and the directional difference between the leading-edge chamber ($\overline{Ch_{right}}$) and following-edge chamber ($\overline{Ch_{left}}$). % from the  and the directional differences in $|STFT_{30}|$. 

The first transition is the initial half contact on the textured surfaces (at 1.1 seconds in \cref{fig:slidingAll}); while the event is unnoticeable in the $\overline{Ch_{all}}$ signal, the $|STFT_{30}|$ of the contacting side $\overline{Ch_{right}}$ is greater than $\overline{Ch_{left}}$. This outcome matches expectations from the vertical orifice simulation in \cref{fig:VerticalOrif}. The following contact transitions,  achieving full contact on the textured surface and then full contact on the smooth surface, can be detected by the increasing $\overline{Ch_{all}}$ over time. The ribbed surface provides less suction engagement than the wavy surface, but both are detectable. During large transitions, defined here as when $\overline{Ch_{all}}$ changes magnitude by over half of the full range signal within a short time-window (i.e., wavy half-to-full contact and ribbed-to-smooth contact), the difference in $|STFT_{30}|$ between $\overline{Ch_{right}}$ and $\overline{Ch_{left}}$ shows characteristic positive-negative shifts; the leakage air on the left side flows more vertically at first (\cref{fig:VerticalOrif}) then its direction transitions to horizontal as more of the suction cup seals with the surfaces (\cref{fig:horizontalOrif}). %However, we did not observe this positive-negative shift when the transition is not drastic. %From these test results on the surface transition, we showed that the tactile sensor can detect spatial contact changes by gradual sliding exploration. 

% Using this We can explore without interfering --> find better surface to engage.
% Interesting transition event can be monitored (start to textured, textured to smooth).

% normal force range was ~ XX newton 

\subsubsection{\textbf{Exploring surface normal of curved objects}}

\begin{figure}[tbp!]
	\centering
	\begin{subfigure}[h]{1\linewidth}
	\centering
	\includegraphics[width=\linewidth]{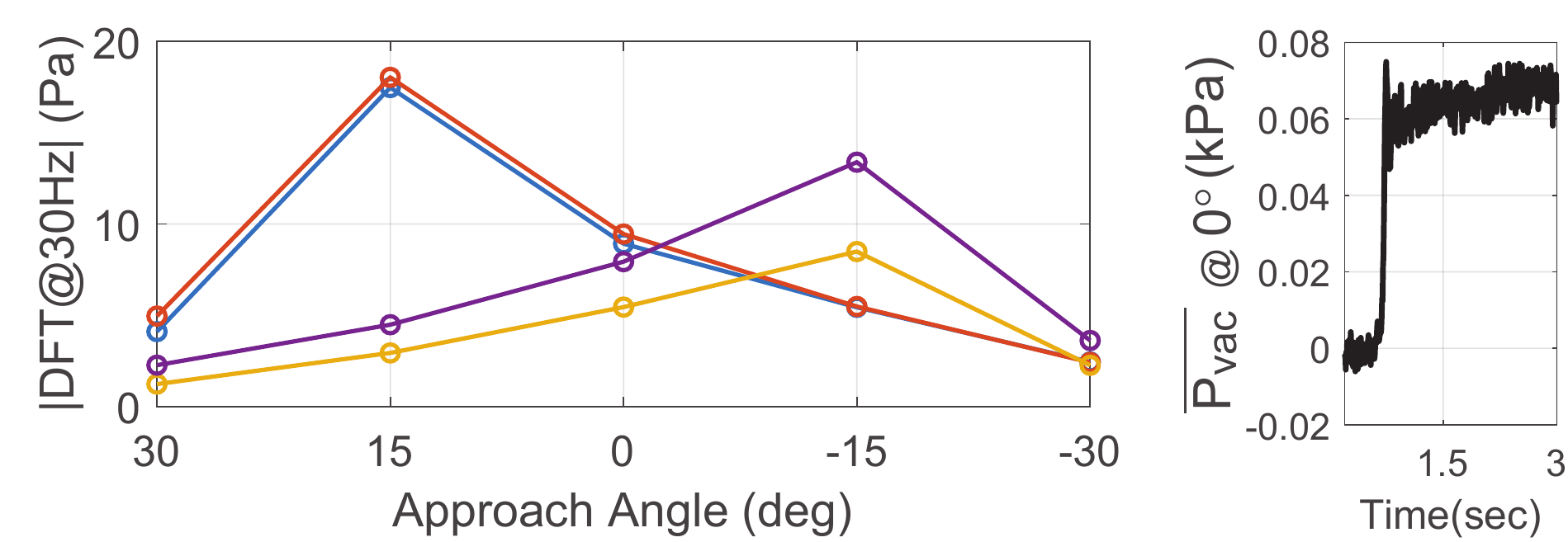}
	\caption{R 8mm, 1N normal force, suction fails }
	\label{fig:mid_1}
        \end{subfigure}
\hfill
    \centering
	\begin{subfigure}[h]{1\linewidth}
	\centering
	\includegraphics[width=1\linewidth]{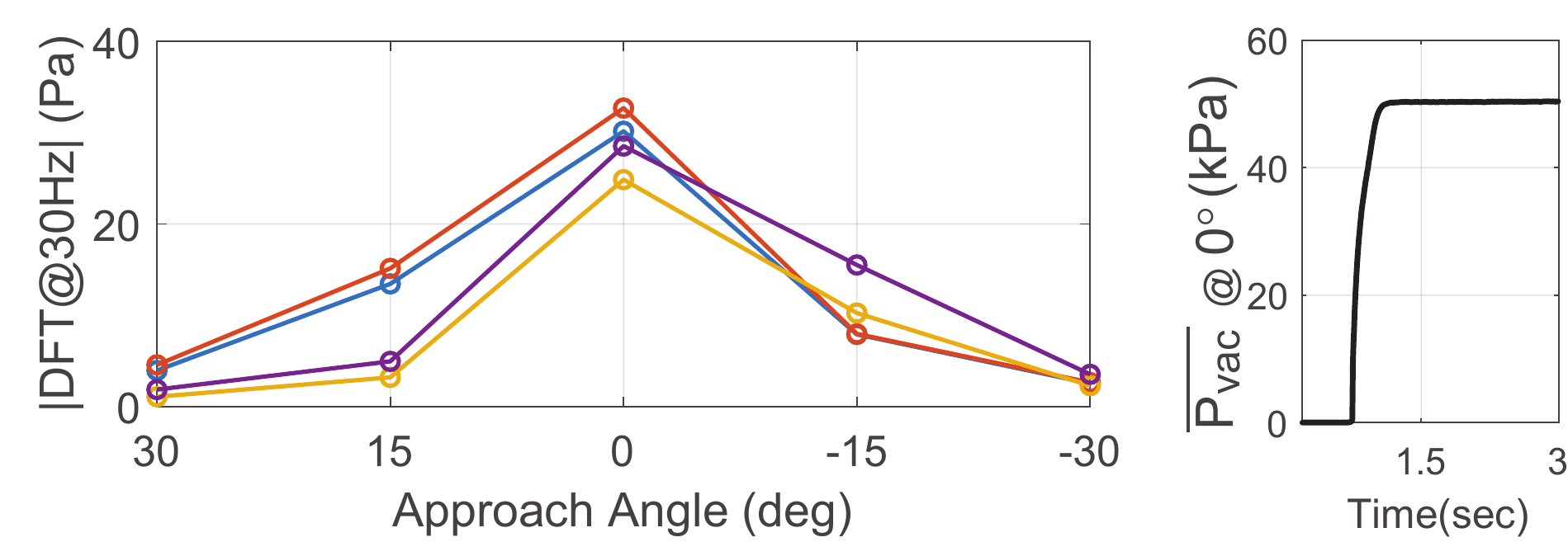}
	\caption{R 11mm, 1N normal force, \textbf{suction succeeds} at 0deg}
	\label{fig:large_1}
        \end{subfigure}
\hfill
    \centering
	\begin{subfigure}[h]{1\linewidth}
	\centering
	\includegraphics[width=\linewidth]{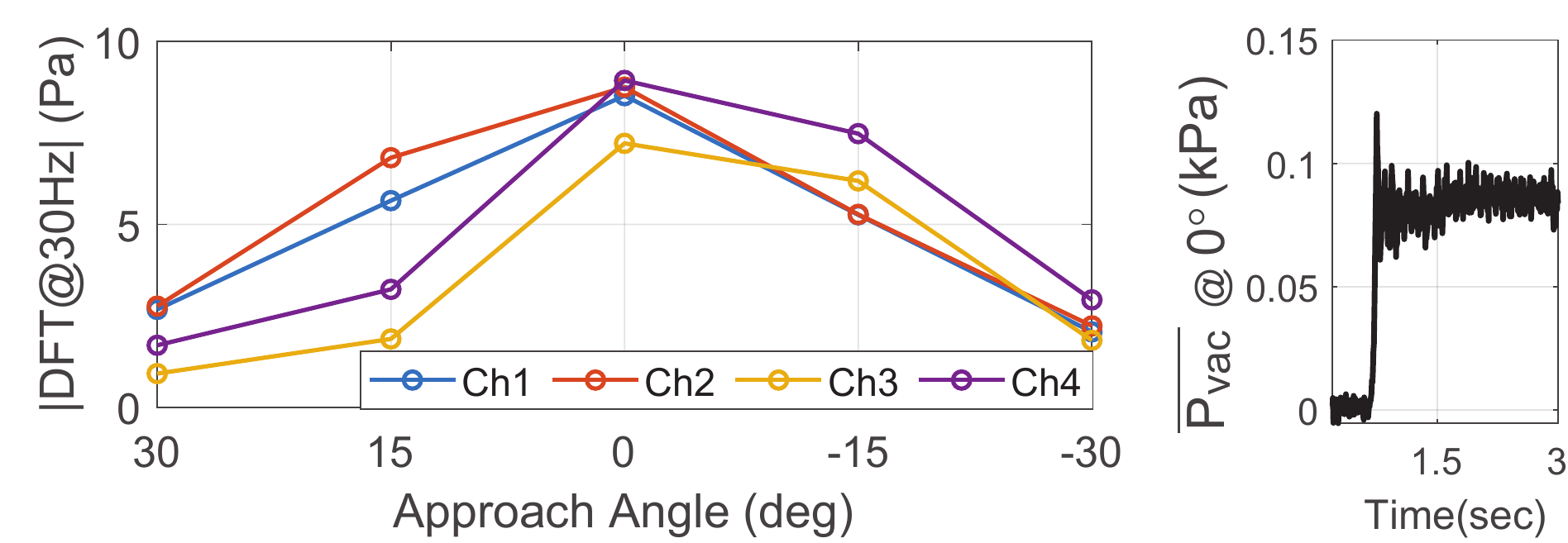}
	\caption{R 11mm, 0.5N normal force, suction fails}
	\label{fig:large_05}
        \end{subfigure}
%\hfill 
%\null
    \vspace{+4pt}
	\caption{Test results of seeking surface normal on the objects in \cref{fig:surfaceNormalSetting}. (Left) Each data-point represents the mean of three trials. Three different curvature-initial normal load pairs all show symmetric trends of $|DFT_{30}|$ about the surface normal (0 deg). The level of $|DFT_{30}|$ at surface normal corresponds to suction grasp success/failure. (Right) mean $|P_{vac}|$ of 4 channels with full vacuum setting at the surface normal (0 deg). \revision{Shown in the right plots, only (b) achieves sufficient vacuum pressure ($\sim 50$kPa) to hold.}}
	\label{fig:surfaceNormalResult}
\vspace{-10pt}
\end{figure}

The surface normals of curved objects are found by the symmetric patterns of $|DFT_{30}|$ over the tested palpation angles. \cref{fig:surfaceNormalResult} shows how this signal varies on the different test objects and with different initial palpation forces. 
%shows the average of three trials. Hannah: I moved this detial to the caption
For all test cases, the $|DFT_{30}|$ of the left channels (Ch1, Ch2) are approximately symmetric to the right channels (Ch3, Ch4) about the surface normal ($\theta=0^\circ$). This trend follows the simulation result in vertical orifice case (\cref{fig:VerticalOrif}), suggesting that the side with less contact produces lower $|DFT_{30}|$. When $\theta=0^\circ$,  $|DFT_{30}|$ in all channel pressures converge to a non-zero value, %the difference between left and right side channels decreases while average $|DFT_{30}|$ is not negligible, 
meaning that leaking is evenly distributed. %normal so the contact between the two sides is comparable. 
$|DFT_{30}|$ of the right side channels are consistently lower than the left channels likely due to both slight experimental misalignment and smart cup manufacturing imperfections. %, meaning that the suction cup axis is slightly misaligned to the ridge of test objects. 
%Despite the misalignment, the symmetric trends of the four-channel outputs are still noticeable.

We posit that the quality of suction contact when $\theta=0^\circ$ %at the surface normal 
can be evaluated by the convergent magnitude of $|DFT_{30}|$. On the object with R=8mm (\cref{fig:mid_1}), $|DFT_{30}|$ at the surface normal ($\theta=0^\circ$) is lower than maximum $|DFT_{30}|$ across all angles testes, meaning that the high curvature of the object prevents complete sealing at any orientation. %is too sharp to engage with our suction cup. 
In contrast, on the lower curvature R=11mm object (\cref{fig:large_1,fig:large_05}) $|DFT_{30}|$ at the surface normal is maximum over all angles tested. %, meaning that the surface normal is better contact than other adjacent angles. 
When the initial palpation force is not sufficient to form a full vacuum seal (\cref{fig:large_05}), the $|DFT_{30}|(\theta=0^\circ)$ is \revision{three orders of magnitude} lower than in suction grasp success (\cref{fig:large_1}). 

%%%%%
%%%%%%
%%%%%%%%%%%%%%%%%%%%%%%%% Surface Normal Result %%%%%%%%%%%%%%%%%%%%%%%%%%%%%
%% Move this paragraph to discussion...
%This proposed method is more useful than using a wrist F/T sensor because it provides close contact information. The directional force/torque measures could be used to search the surface normal of the contact point. However, when the robot needs to explore an object geometry in a cluttered environment, common in a bin of an e-commerce warehouse, the wrist F/T sensor measure could also be affected by other adjacent objects. Moreover, even when the surface normal is found with the F/T sensor, the information whether the contact will form a suction seal is still missing. Therefore, our proposed geometric surface approach should provide more useful information about contact geometry in practice.

\section{Predicting Leakage and Detachment}\label{sec:Detach}

Monitoring leaks in the seal of a suction cup can help in preventing grasp failures during robotic manipulation, especially of unknown objects that may have %shifting mass or 
variable surface properties. %In recent work on suction cup manipulation using extrinsic dexterity\cite{cheng2019manipulation} and fast dynamic motions\cite{Pham2019}, the controller assumed known inertia properties of an object to optimize the physics of object motions. However, in practice, the inertia of an object is rarely known and an object may contain moving materials (e.g. water in a bottle) with difficult-to-model dynamics.
We hypothesize that it may be possible to detect and localize breaks in the suction seal early enough to avoid total astrictive grasp failure. % or before it occurs, 
%adaptive controller may recover by applying different wrenches accordingly. 
As a first step toward an adaptive controller capable of such behaviors, we examine the capability of the gripper to monitor spatial vacuum seal states and localize the leakage airflow in both current and future states in a controlled experiment.

\subsection{Dataset}
We estimate the contact seal states around the perimeter of the suction cup while collecting time-series tactile sensing and wrist F/T sensor data as the suction cup is detached from a smooth flat plate. The ground-truth contact seal states is measured using a Frustrated Total Internal Reflection (FTIR) setup, as in Li et al.~\cite{li2020milliscale}. A 13mm thick smooth acrylic plate is illuminated from the side with red color filtered LED strips. When the suction cup engages with the acrylic, the contact regions appear bright red to a high speed digital camera (240 FPS) that records images from the opposite side.

\begin{figure}[tbp!]
\centering 
	%\vspace{-10pt}
	%\begin{subfigure}[h]{0.5\textwidth}
	%\centering
	\includegraphics[width=0.6\linewidth]{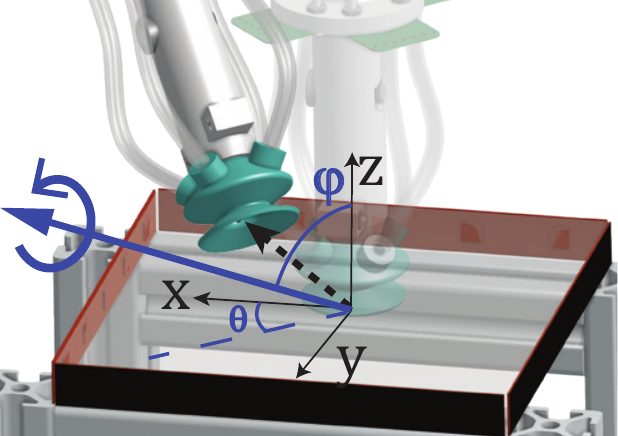}
	%\end{subfigure}
    \vspace{+5pt}
	\caption{Frustrated Total Internal Reflection(FTIR) test setup and robot motion with an example twist. Dashed arrow is a linear velocity and the purple arrow is an axis of angular velocity.}
	\label{fig:FTIR_setup}
	\vspace{-10pt}
\end{figure}

The suction cup is forcefully detached from the acrylic substrate under maximum vacuum for 740 separate trials (\cref{fig:FTIR_setup}). The UR-10 robot arm applies 450g of %the astrictive grasp is initiated such that it is pulling away from the surface with 450g. %
initial vacuum suction loading away from the substrate. % equivalent to 450g. 
Then the robot applies a twist to the end-effector until it detaches. The angular speeds are set to 30\degree$\pm$5\degree /sec and the axis of rotation is set by $\varphi$ in [0\degree, 180\degree] with 22.5\degree increments and $\theta$ in [0\degree, 330\degree] with 30\degree increments. Each pair of $\varphi$ and $\theta$ is tested at least six times, and  a uniformly distributed perturbation in [-0.5, 0.5]$\times$ increment is added to $\varphi$ and $\theta$. The translation velocity is chosen randomly between [-1, 1] cm/sec for (x,y) axis and [0.6, 2] cm/sec for (z) axis. %In this experiment, we use fully powered vacuum pressure, assuming a manipulation of an object with fully engaged suction grasping. 
At the end of each detachment, the high-speed video is synchronized with other sensor data by blinking the FTIR LED for timestamping. % and aligning timestamps. 

% We applied randomized 740 wrenches to detach the suction cup from the acrylic substrate (\cref{fig:FTIR_setup}). Using the UR-10 robot arm, we set the initial suction loading to be equivalent to 450g. Then the robot translates the end-effector to a randomly chosen position of (x,y) $\in$ [-1cm, 1cm] and (z) $\in$ [1cm, 3cm] while rotating it around a randomized quaternion rotation axis. The angle of rotation ($\gamma$) was fixed to 45deg for simplicity and the rotational speed was approximately 30 deg/sec. In this experiment, we used fully powered vacuum pressure, assuming a manipulation of an object with fully engaged suction grasping.
%%%%%
%%%%%%
%%%%%
\begin{figure}[tbp!]
\centering
	%\vspace{-10pt}
	%\begin{subfigure}[h]{0.5\textwidth}
	%\centering
	\includegraphics[width=\linewidth]{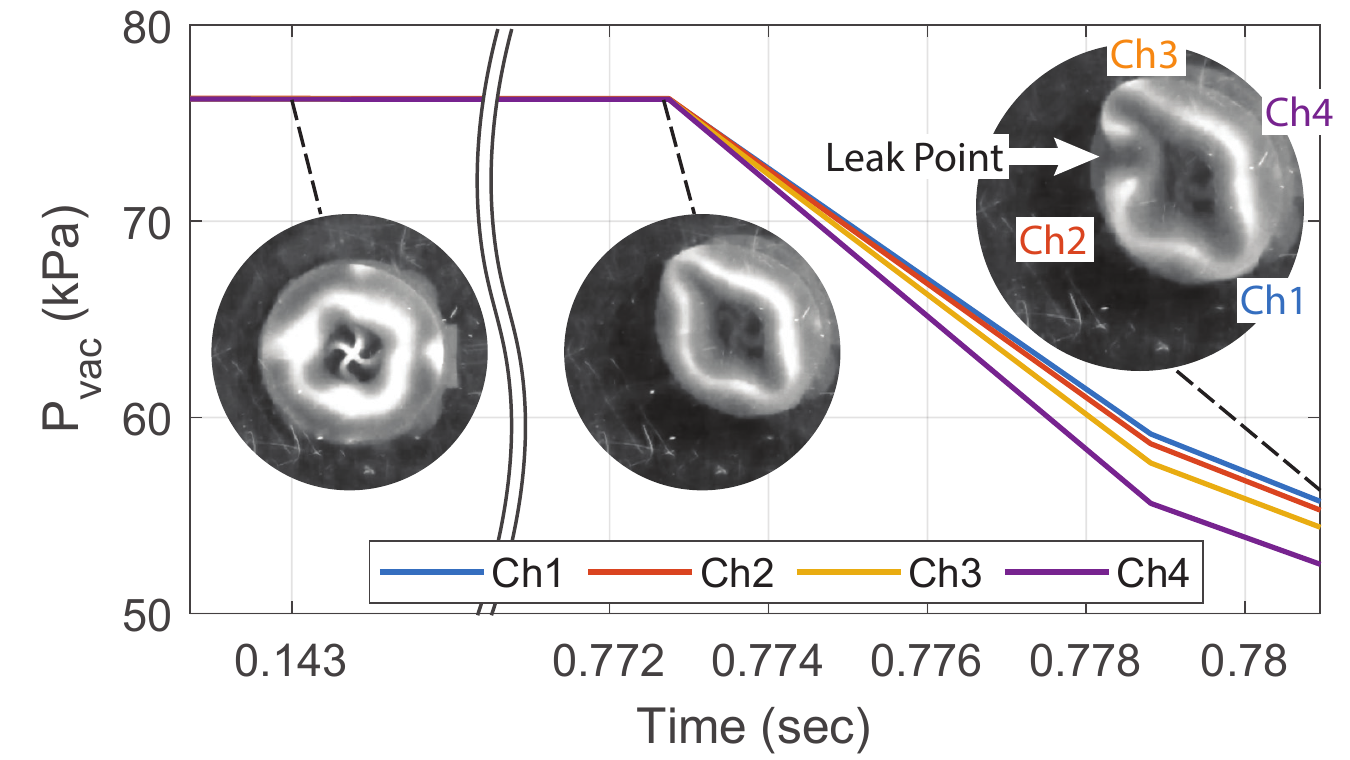}
	%\end{subfigure}
    %\vspace{+1pt}
	\caption{Vacuum pressure measured at each chamber during detaching motion. Insets are gray-scaled image frames from the FTIR sensor. %In the right inset, the leak point is in the yellow sector of which contact label is 0.53.
	}
	\label{fig:FTIR_example}
	\vspace{-10pt}
\end{figure}
%%%%%%
%%%%%
\begin{figure}[tbp!]
\centering
	%\vspace{-10pt}
	%\begin{subfigure}[h]{0.5\textwidth}
	%\centering
	\includegraphics[width=0.75\linewidth]{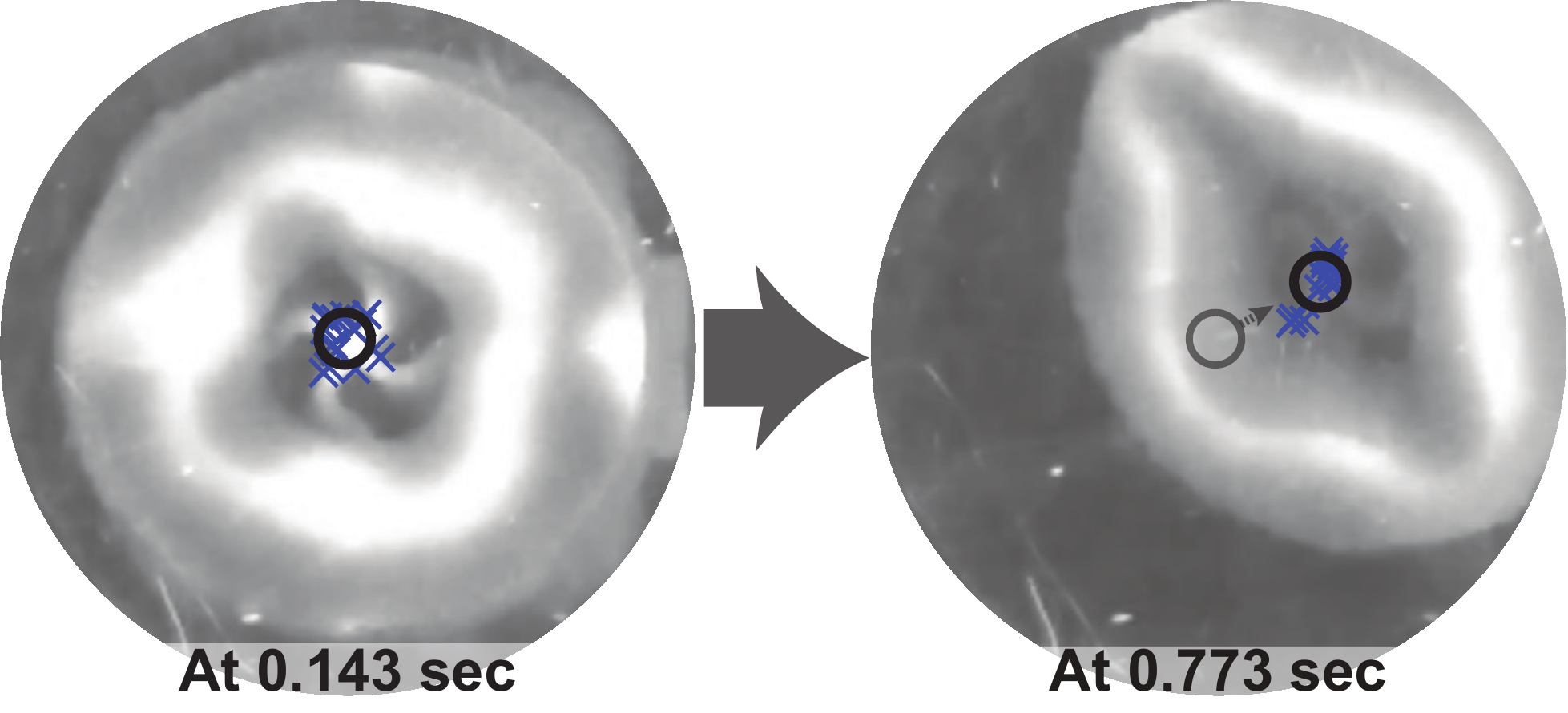}
	%\end{subfigure}
    %\vspace{+1pt}
	\caption{Photo of center point tracking in the image sequence. The characteristic points (blue `X' in images) around the center of suction cup are tracked throughout the video sequence. The center point (black circle) is the mean coordinate of all tracked points.
	}
	\label{fig:FTIR_PointTrack}
	\vspace{-10pt}
\end{figure}
%%%%%%
%%%%%
\begin{figure}[tbp!]
\centering
	%\vspace{-10pt}
	%\begin{subfigure}[h]{0.5\textwidth}
	%\centering
	\includegraphics[width=0.9\linewidth]{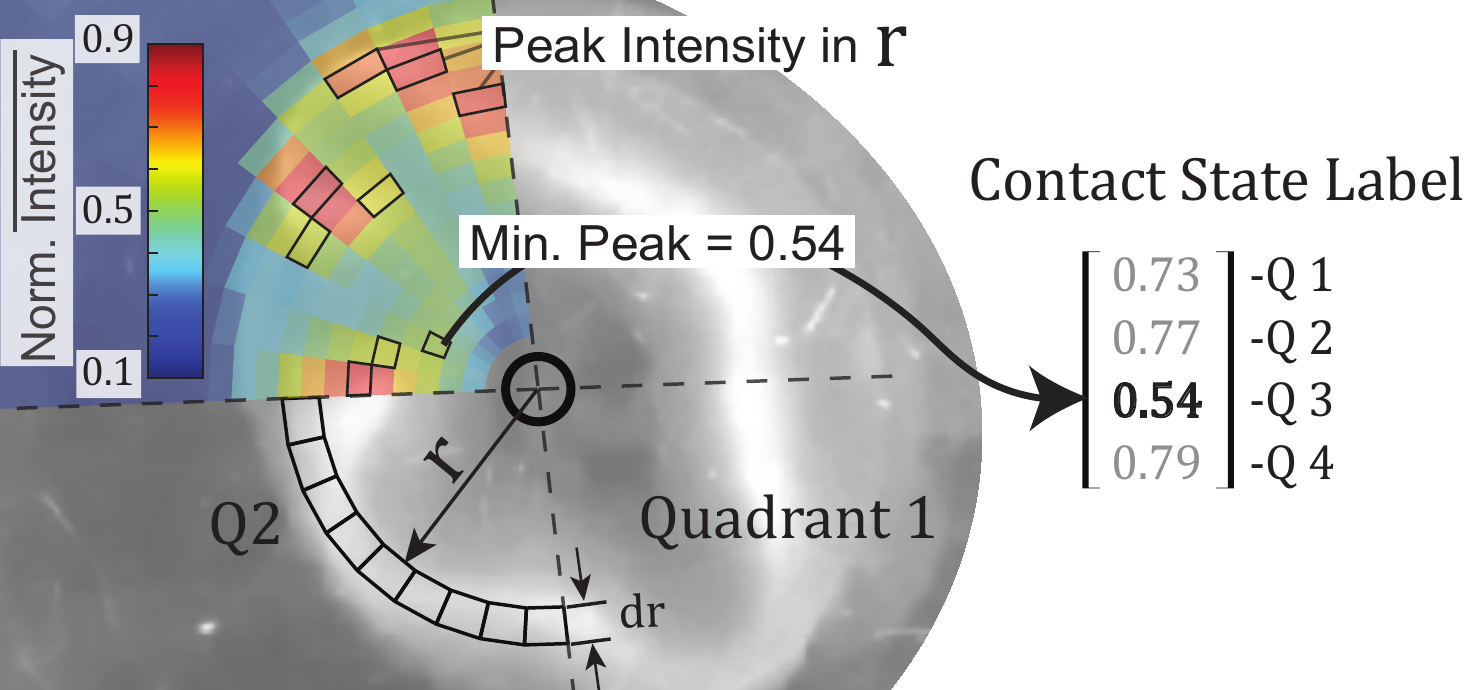}
	%\end{subfigure}
    \vspace{+0.5pt}
	\caption{Process of obtaining contact state label from a single frame. (Left) the quadrants are defined around the tracked center point. For each quadrant, the averages of normalized intensity in a small radial band ($dr = 16$ pix) of subsectors (9 per quadrant) are evaluated along the radial direction ($r$). Minimum peak among nine subsectors is considered the contact state. (Right) contact label computed for all quadrants results in a $4\times1$ vector.
	}
% 	\caption{Process of obtaining contact state label from a single frame. (Top left) the 12 sectors are defined around the tracked center point. (Bottom left) For each sector, the averages of normalized intensity in a small radial band ($dr = 16$ pix) of subsectors (A,B,C) are evaluated along the radial direction ($r$). The minimum peak among three subsectors is considered the contact state. (Right) the contact label is computed for all sectors, resulting in a $12\times1$ vector.
% 	}
	\label{fig:FTIR_dataset}
	\vspace{-10pt}
\end{figure}

%%%%%%

\begin{table}[b!]
    \renewcommand{\arraystretch}{0.7}
	\centering
	\vspace{-5pt}
	\begin{tabu} to \linewidth {X[1.5c]X[c]X[c]X[c]X[c]}
            % &  \multicolumn{4}{c}{\textbf{Break occurrence rate}}\\
        \toprule
		  \textbf{Threshold (\textit{th})} & \textbf{Q1} & \textbf{Q2} & \textbf{Q3} & \textbf{Q4}\\\midrule
		  0.5 & 0.50 & 0.65 & 0.6 & 0.40\\
		  0.6 & 0.34 & 0.46 & 0.59 & 0.44\\
		  0.7 & 0.19 & 0.14 & \textbf{0.87} & 0.08\\ \bottomrule
	\end{tabu}
% 	\vspace{1pt}
	\caption{Break occurrence rate of each quadrant where the contact label first decreases below thresholds (0.5-0.7) in 740 trials. Multiple quadrants can break the threshold at the same time, making the rate sum $\geq$1.0.}
	\label{tab:breakPointDistribution}
		%\vspace{-10pt}
\end{table}

\begin{table*}[th!]
    \renewcommand{\arraystretch}{0.7}
    \setlength{\tabcolsep}{0.7pt}
	\centering
	\begin{tabu} to 0.49\linewidth {X[1.2l]X[c]X[0.8c]X[0.8c]X[0.8c]X[c]X[c]X[1.3c]}
	\toprule
		\revision{\Large{\textit{\textbf{XGBoost}}}}&  & \multicolumn{3}{c}{\textbf{BQA} $\bm{\uparrow}$} & \multicolumn{3}{c}{\textbf{MBTE (ms)}} \\
		\cmidrule(lr){3-5}\cmidrule(lr){6-8}
		\textbf{Model} & \textbf{MSE} $\bm{\downarrow}$ & $th=0.5$ & $th=0.6$ & $th=0.7$ & $th=0.5$ & $th=0.6$ & $th=0.7$ \\\midrule
        FT Only & $0.017$ & $35\%$ & $26\%$ & $40\%$ & $\bm{-6 \pm 18}$ &  $-6 \pm 12$ & $-30 \pm 30$ \\
        Vac Only & $0.018$ & $\bm{44\%}$ & $33\%$ & $\bm{70\%}$ & $-6 \pm 24$ & $\bm{-6 \pm 6}$ & $\bm{18 \pm 60}$ \\
		FT+Vac & $\bm{0.012}$ & $39\%$ & $\bm{35\%}$ & $67\%$ & $-6 \pm24$ & $\bm{-6 \pm 6}$ & $-30 \pm 24$ \\\bottomrule
	\end{tabu}
	\hfill
	\begin{tabu} to 0.49\linewidth {X[1.2l]X[c]X[0.8c]X[0.8c]X[0.8c]X[c]X[c]X[1.3c]}
	\toprule
		\Large{\textit{\textbf{LSTM}}}&  & \multicolumn{3}{c}{\textbf{BQA} $\bm{\uparrow}$} & \multicolumn{3}{c}{\textbf{MBTE (ms)}} \\
		\cmidrule(lr){3-5}\cmidrule(lr){6-8}
		\textbf{Model} & \textbf{MSE} $\bm{\downarrow}$ & $th=0.5$ & $th=0.6$ & $th=0.7$ & $th=0.5$ & $th=0.6$ & $th=0.7$ \\\midrule
        FT Only & $\bm{0.003}$ & $86\%$ & $82\%$ & $\bm{88\%}$ & $\bm{6 \pm 18}$ &  $\bm{0 \pm 24}$ & $\bm{0 \pm 44}$ \\
        Vac Only & $0.006$ & $80\%$ & $\bm{97\%}$ & $82\%$ & $18 \pm 18$ & $24 \pm 12$ & $-24 \pm 138$ \\
		FT+Vac & $0.006$ & $\bm{95\%}$ & $82\%$ & $82\%$ & $24 \pm 6$ & $24 \pm 12$ & $-12 \pm 144$ \\\bottomrule
	\end{tabu}
	
	\caption{\textbf{Learning to predict suction failure.} Mean squared error (MSE) of predicted contact state, break quadrant accuracy (BQA), and median break time error (MBTE) (in timesteps with interquartile range shown) for models trained on vacuum pressure, force/torque data, or both. BQA and MBTE are reported for different failure cutoff thresholds $th=0.5,0.6,$ and $0.7$. Positive MBTE values indicate the model is overly conservative, while negative MBTE values indicate the model does not predict the breakage until it is imminent; in practice, an overly conservative model such as FT+Vac may be more beneficial to a system that must quickly react to an imminent breakage. %All models learn to localize the first leakage point in both space and time for lower thresholds, but accuracy decreases as the threshold increases.
	\revision{LSTM outperforms XGBoost in all metrics.}}
	\vspace{-10pt}
	\label{tab:lstmresults}
\end{table*}

An example sequence is shown in \cref{fig:FTIR_example}. As the suction cup deforms under the detaching twist, the vacuum seal (bright ring) also deforms and then leaks. The $P_{vac}$ measure in each chamber does not vary while the vacuum seal remains, but lowers at the onset of the leak. The difference in $P_{vac}$ among chambers shows expected trends based on the horizontal leakage CFD simulation in \cref{fig:horizontalOrif}. The example in \cref{fig:FTIR_example} shows the least $P_{vac}$ in the chamber opposite to the leak point; the leak occurs near `Ch2' and the pressure drops most rapidly for `Ch4'. The large deformations of this suction cup are a consequence of using soft rubber to ease the fabrication process, described in \cref{sec:DesignFab}. To estimate local vacuum seal states in this highly deformable suction cup, we trained an artificial neural network using $P_{vac}$ signals and wrist F/T sensor data as inputs. 

%\todo{Although the pressure measures provide some estimation of leaking points, large deformation of the suction cup may make it difficult to monitor accurate contact states. Therefore, we also include the wrist F/T sensor data for deep learning training, assuming the measured wrench provides more global deformation and the $P_{vac}$ can provides local vacuum seal details.}

For training labels, we process the video data to obtain time-series contact states. We segment the suction cup into 4 quadrants around the center, and we define the contact state label of a single video frame as a vector of 4 elements that represents the contact of each sector. Because the center point moves, as shown in \cref{fig:FTIR_example}, we track the center using the Kanade-Lucas-Tomasi (KLT) algorithm \cite{shi1994good} featured in MATLAB. As shown in \cref{fig:FTIR_PointTrack}, the characteristic points ($\sim$15 points) around the center are selected from the first contact image by using minimum eigenvalue corner detection algorithm \cite{shi1994good}. Then the KLT algorithm tracks the points in the video frame sequence. The mean of the coordinates of tracked points are assumed to be the center point. If the algorithm losses track of more than half of the points, the center point is extrapolated from previous center trajectory. 

Then the contact label is computed around the tracked center point as shown in \cref{fig:FTIR_dataset}. The four quadrants are redefined every frame from the tracked center point and the interpolated end-effector orientation that is sampled at 10Hz. We assume that the contact state of each sector is dictated by the weakest, or darkest, contact points along the suction seal. For each sector, we use the normalized image intensity values as a metric of contact states, and compute the average intensity in a small subsector band (dr = 16 pixels) along the radial direction (\cref{fig:FTIR_dataset}). The minimum peak of mean intensity values among the nine subsectors is assumed %to represent the weakest contact seal in this sector, thus we used it 
as that quadrant's contact label. \cref{tab:breakPointDistribution} shows that catastrophic vacuum seal break, defined here as dropping below a 50\% or 60\% contact state threshold ($th = 0.5$ or $0.6$), occurs evenly throughout the four quadrants. Moderate seal weakening ($th = 0.7$) occurs more at a particular quadrant, which may be caused by mechanical bias from the fabrication process.
% \todo{Missing: 1 sentence describing how much the `leak point' sector varies throughout the twists tested... is it always in the same sector for this dataset or does it span a wide range of sectors?}

The time series of the contact state labels in all 4 quadrants (size: $t \times4$) are the output data, and the time series of 4-channel $P_{vac}$ and 6DOF F/T sensor data (size: $t \times (4+6)$) are the input data of the deep learning training. All timestamps of F/T sensor data and contact state labels are interpolated to match the $P_{vac}$ timestamps.

\begin{figure}[tbp!]
\centering
	\vspace{-10pt}
% 	\begin{subfigure}[h]{0.5\textwidth}
	\centering	\includegraphics[width=0.9\linewidth]{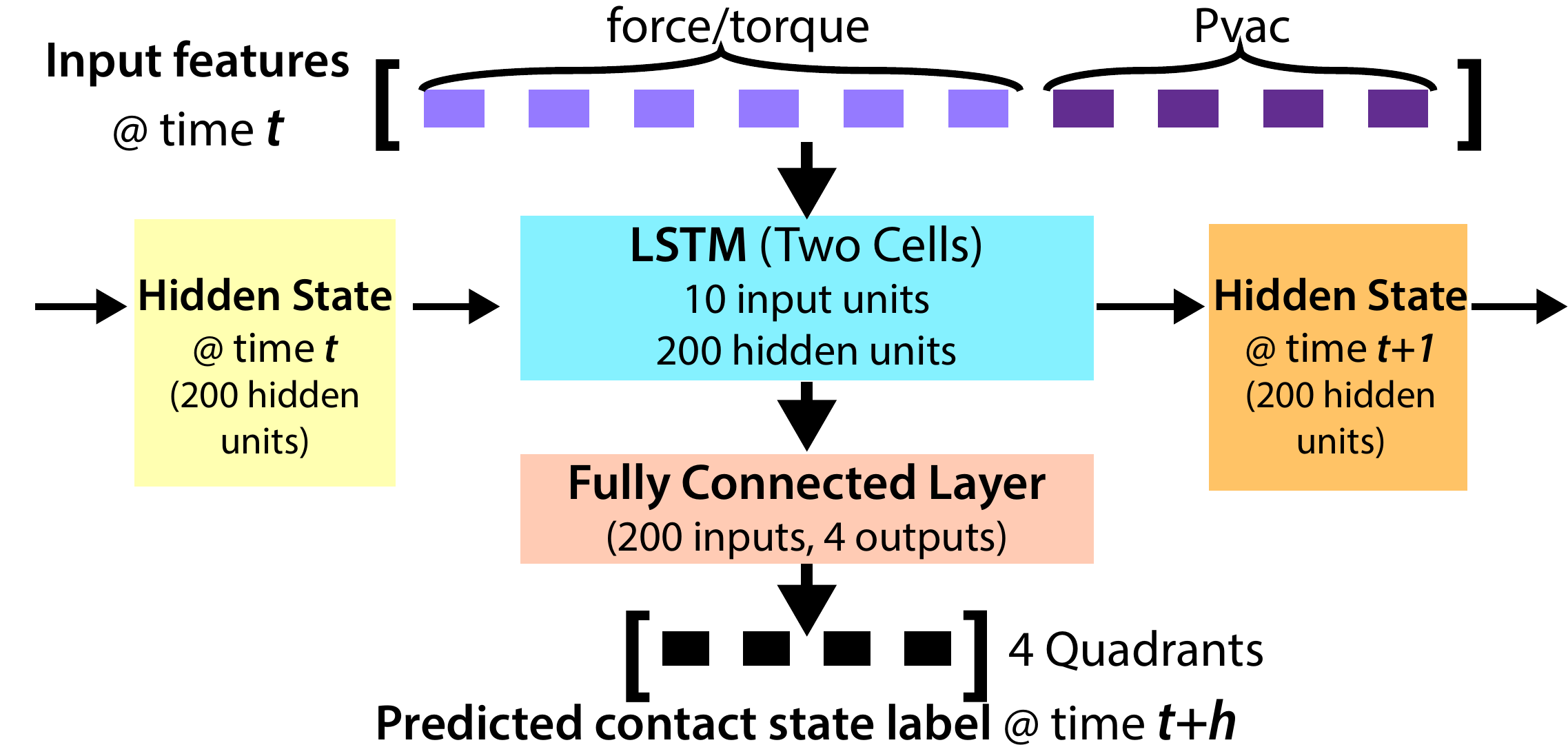}
% 	\end{subfigure}
    \vspace{+1pt}
	\caption{Architecture of the deep learning model used to learn contact state. Takes force, torque, and pressure features at timestep $t$ and the previous step's hidden cell as input, and outputs a new hidden state along with features that are passed through a fully connected layer to produce the contact state scores for each section of the suction cup.
	}
	\label{fig:architecture}
	\vspace{-10pt}
\end{figure}

\begin{comment}
\subsection{LSTM Training}
To predict both the current seal state as well as future seal states, we trained a long short term memory (LSTM) network~\cite{hochreiter1997long} on the given input data. The architecture is represented in \cref{fig:architecture}. Specifically, we wish to predict the contact state of a suction cup at time $t+h$ given the F/T and pressure sensor data from time $[0, t]$, where $h$ is the prediction horizon. We aim to predict $h$ ms into the future based on current sensor data as this would allow for preventative measures to be executed before suction failure occurs on the actual machine if this model were employed in a system. We use two LSTM cells with 10 input features and 200 hidden layer units, followed by a fully-connected layer with 200 input features and 4 output features. The memory unit of the LSTM allows the model to learn key features of time series data, and 200 hidden units allows for the model to develop a robust representation of the input data sequence while still being fast to train. At prediction step $t'$, sensor data for time $t=t'$ is passed into the LSTM and contact state scores are produced for time $t=t'+h$. The four contact state outputs correspond to one quadrant of the suction cup. We trained and validated on 592 detachment sequences, leaving 148 sequences as a test set. We train the network for 100 epochs using the Adam optimizer and a learning rate of 0.001. The LSTM training takes about 5 minutes on an NVIDIA Titan X GPU.
\end{comment}

\revision{
\subsection{Training}
We wish to predict the contact state of a suction cup at time $t+h$ given the F/T and pressure sensor data from time $[0, t]$, where $h$ is the prediction horizon. We aim to predict $h$ ms into the future based on current sensor data as this would allow for preventative measures to be executed before suction failure occurs on the actual machine if this model were employed in a system. 
% We comment that naive time series forecasting methods (e.g., ARIMAX) is not well-suited because historical ground-truth contact labels are unavailable for autoregression during prediction time. Instead, 
We compare two methods: a sliding-window-based regression model and a long short term memory (LSTM) network~\cite{hochreiter1997long}. For sliding-window regression, we experiment with different sliding window lengths $l$ and regression methods. We find that $l = 60$~ms combined with XGBoost~\cite{chen2016xgboost} works relatively well. At each time step $t'$, input features between time $t'-l$ and $t'$ are passed into the regression model to predict the contact state scores at time $t=t'+h$. We set the maximum depth of a tree to be 5 and the learning rate to be 1 for XGBoost training.
The LSTM architecture is represented in \cref{fig:architecture}. We use two LSTM cells with 10 input features and 200 hidden layer units, followed by a fully-connected layer with 200 input features and 4 output features. The memory unit of the LSTM allows the model to learn key features of time series data, and 200 hidden units allows for the model to develop a robust representation of the input data sequence while still being fast to train. At prediction step $t'$, sensor data for time $t=t'$ is passed into the LSTM and contact state scores are produced for time $t=t'+h$. The four contact state outputs correspond to one quadrant of the suction cup. For both models, we train and validate on 592 detachment sequences, leaving 148 sequences as a test set. We train the LSTM network for 100 epochs using the Adam optimizer and a learning rate of 0.001. The training takes about 5 minutes on an NVIDIA Titan X GPU.
}

\subsection{Results}
\begin{comment}
To gauge the ability of the LSTM to predict both the seal state over time as well as its ability to predict and localize impending seal breakage, we use three evaluation metrics: 1) mean squared error (MSE), which measures the ability of the LSTM to track the contact seal state across the entire sequence, 2) break quadrant accuracy (BQA), which measures the ability of the LSTM to localize the quadrant of the first seal break across each sequence, and 3) median break time error (MBTE), which measures the ability of the LSTM to localize the seal breakage in time.
\end{comment}

To gauge the ability of the \revision{two models} to predict both the seal state over time as well as its ability to predict impending seal breakage, we introduce three evaluation metrics: 1) mean squared error (MSE), which measures the ability of the \revision{models} to track the contact seal state across the entire sequence, 2) break quadrant accuracy (BQA), which measures the ability of the \revision{models} to localize the quadrant of the first seal break across each sequence, and 3) median break time error (MBTE), which measures the ability of the \revision{models} to localize the seal breakage in time.

% \begin{align*}
%     MSE &= \frac{1}{n} \sum_{i=1}^n (y_i - \hat{y}_i)^2 \\
%     BQA &= |\argmin_s \argmax_i \hat{\mathbf{y}} \ \cap \ \argmin_s \argmax_i \mathbf{y}| > 0 \\
%     BTE &= \text{median}\;(y_i- \hat{y}_i)
% \end{align*}

Mean squared error is calculated between the normalized predicted and ground-truth contact values, averaged over the length of the sequence. Break quadrant accuracy is calculated by comparing the first quadrant to dip below a contact state threshold $th$ in each of the ground-truth and predicted series. Median break time error (with the interquartile range also indicated in the table) is calculated by averaging the error in time steps between the first time the first quadrant drops below $th$ in the predicted and ground-truth series, respectively. The latter two metrics are particularly useful for an adaptive controller, which must robustly identify catastrophic seal failure before it occurs. 

\begin{figure}[tbp!]
\centering
% 	\vspace{-10pt}
	%\begin{subfigure}[h]{0.5\textwidth}
	%\centering
	\includegraphics[width=0.8\linewidth]{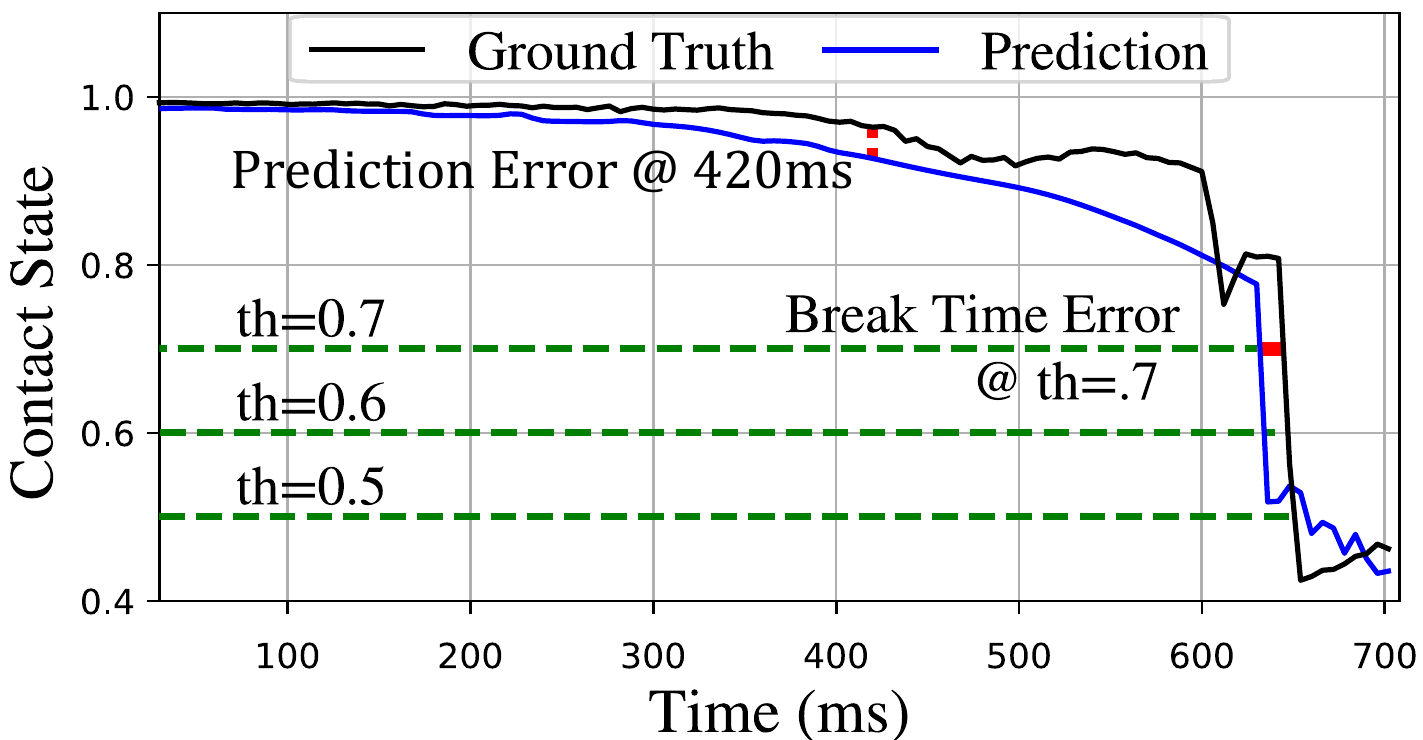}
	%\end{subfigure}
    %\vspace{+1pt}
	\caption{LSTM Contact state prediction results for one seal quadrant of a previously unseen trial. Model prediction delay is set to $h=30$ ms, so the contact state prediction at time $t$ on the graph is produced using sensor data through time $t-h$. Figure labels $y=0.5$, 0.6, and 0.7 thresholds used to calculate Median Break Time Error, which is also labeled at $th=0.7$. Raw error used to calculate the MSE is also labeled at $t=420$ ms.}
	\label{fig:sealPrediction}
	\vspace{-10pt}
\end{figure}

Table~\ref{tab:lstmresults} shows the results of training the \revision{XGBoost and the LSTM} network using just F/T sensor data (FT Only), just vacuum sensor data (Vac Only), and using both inputs (FT+Vac). In all cases, the \revision{LSTM} network learns to predict the contact state with low MSE, and the FT+Vac model can additionally learn to robustly predict the first quadrant that will leak below a threshold of $0.5$. \revision{In contrast, the MSE for XGBoost is about three times as large as that using LSTM, and we find that its break quadrant prediction is only about half as accurate as that using LSTM.}
%\revision{For both algorithms}, as the failure cutoff threshold increases, the predictions are less accurate in both space and time. 
\revision{While the BQA of sliding-window regression drops significantly as the failure cutoff threshold decreases, LSTM's break quadrant prediction becomes more accurate.}
This suggests that the LSTM network can accurately identify impending catastrophic failures, but may not be able to completely model stretching or bending effects that can cause smaller amount of leakage. The median break time error results \revision{for LSTM} suggest that the force/torque data alone can accurately identify the breakage time, but adding the vacuum pressure inputs causes the model to predict a breakage more conservatively. Since positive MBTE values indicate an overly conservative prediction (i.e., the model predicts a breakage will happen in 30~ms when it will actually occur in 54~ms), this conservative nature may actually benefit the system in practice, as it can react with more time before the breakage. In contrast, the F/T Only model can sometimes be too optimistic (negative MBTE).

% The median break time error shown indicates that, on average, the model predicts suction failure 24 ms time steps early with an error of 1 or 2 time steps, for thresholds of $0.5$ and $0.6$ respectively, when using both F/T and vacuum tactile inputs. Model performance with respect to this metric indicate that the vacuum sensor is able to provide information that allows a deep learning model to better predict suction failure ahead of it occurring on the actual machine, which is illustrated by the positive median break time errors the FT+Vac and Vac Only models achieve for $th=0.5$ and $th=0.6$. Data indicates that, on average for these thresholds, the FT Only model predicts failure later than the other models and has a higher probability of predicting a failure after it has already occurred in the ground truth data. 
Ablation experiments of the LSTM on the prediction horizon parameter, $h$, suggest a linear relationship between the MSE and $h$, as the MSE increases by approximately 0.002 for every 60 ms $h$ is increased by over a range of $30\leq h\leq 330$. Data indicates that higher horizon values correspond to large, negative BTE scores, such that predictions become inaccurate and late, on average, as the horizon grows.
\section{Discussion}
\label{sec:conclusion}

% \han{Hannah: I suggest we use this section for critical discussion of the results and future work, and reduce the amount to which we are simply repeating what we already said earlier.}

A single-bellows suction cup gripper can examine object surfaces by haptic exploration and monitor detachment details of the vacuum seal. The four-chamber cup design %We designed wall structures inside the suction cup that separate airflows into four chambers. The 
generates differential airflows between the chambers, measured with remote pressure transducers such that sensitive electronics are not exposed to physical damage or fatigue. 
We find that lowered vacuum pressure facilitates gentle interactions with objects. 
% and we detect local suction contact events using them. 
%The use cases analyzed in this work demonstrate the complexity and richness of these signals, such that this sensor is well suited to use with data-driven analysis methods in robotic applications. %We simulated the pressure difference in two exemplary contact cases (vertical and horizontal airflow), and the two cases show different pressure distributions throughout the four chambers. 
%A primary utility of this smart suction cup is in haptic exploration through touch. %. Using the pressure differences and overall pressure, we demonstrate a haptic exploration of a suction cup gripper to 
During haptic exploration, this smart suction cup responds to surface textures, transitions between different surfaces, surface normal and local curvature of touched surfaces. %We expect these parameters to inform new grasp planning algorithms. %We find that lowered vacuum pressure facilitates gentle exploratory interaction with an object. %Using this method, the suction gripper slides over a surface to identify the transition of surface textures, and we 
The next step is to further characterize how these gentle and exploratory haptic signals correlate with strong astrictive grasp wrench limits, so that the gripper can apply more versatile planning and adaptive control on a wide variety of objects. %on various surfaces as a means to further inform grasp planning and control. %  graspability  of the surface normal on curved objects. 

This tactile sensor detects spatial and temporal details of suction cup detachments when applied to an LSTM network. In these trials, the suction cup sensor ($<$\$80 for the raw components) predicts the quadrant and time of initial breaks with comparable accuracy as using a wrist force/torque sensor (ATI Axia80, $\sim$\$3,000). However, it is not our intention to create an equivalent to the common wrist load-cell.
%The network could predict the catastrophic failure within \todo{xx ms-need to double check with Mike}, which potentially be useful for the adaptive manipulation control that prevents failure. Hannah: This feels more like a result... report earlier in the paper?
While our results highlight the smart suction cup's sensitivity to external loads through deformation,  we only test on a smooth, flat, fixed acrylic plate. % from a fixed smooth acrylic plate. %In unstructured tasks, where a variety of unconstrained objects are dynamically manipulated, flow characteristics and astrictive grasp wrench limits will change with object surface properties (e.g., texture, porosity, rugosity, etc.). 
Future work will characterize the effect of unknown, variable surface properties on the prediction of leaks, which we expect to accentuate the importance of internal suction cup flow sensing over the measurement of external grasp forces alone. \revision{Then, we will explore responsive control of the smart suction cup to prevent impending grasp failure.}
%Future work will seek to characterize how the measurement of internal suction cup flow provides a more direct measure of imminent grasp failure than the measurement of external grasp forces alone. 

% Astrictive grasp wrench limits are influenced by porosity, texture and other non-idealities of the surface and cannot necessarily be characterized by a wrist load cell and visual sensors alone.}

%Future work 
% - estimate the maximum pull force from lowered vacuum exploration mode.
% - adaptive control to prevent detach failure.
% W

\section{Acknowledgments}
T.M.Huh was supported by the University of California at Berkeley Embodied Dexterity Group. M. Danielczuk was supported by the National Science Foundation Graduate Research Fellowship Program under Grant No. 1752814. The work of M. Li was
supported by the NASA Space Technology Research Fellowship, under Grant
\#80NSSC19K1166.

\bibliographystyle{IEEEtran}

\balance  % We need this line for the correct ordering of reference. ( this line is in "balance" Package

\bibliography{IEEEabrv,Biblio}   % Reference are in "Biblio.bib" file.

\begin{thebibliography}{10}
\providecommand{\url}[1]{#1}
\csname url@samestyle\endcsname
\providecommand{\newblock}{\relax}
\providecommand{\bibinfo}[2]{#2}
\providecommand{\BIBentrySTDinterwordspacing}{\spaceskip=0pt\relax}
\providecommand{\BIBentryALTinterwordstretchfactor}{4}
\providecommand{\BIBentryALTinterwordspacing}{\spaceskip=\fontdimen2\font plus
\BIBentryALTinterwordstretchfactor\fontdimen3\font minus
  \fontdimen4\font\relax}
\providecommand{\BIBforeignlanguage}[2]{{%
\expandafter\ifx\csname l@#1\endcsname\relax
\typeout{** WARNING: IEEEtran.bst: No hyphenation pattern has been}%
\typeout{** loaded for the language `#1'. Using the pattern for}%
\typeout{** the default language instead.}%
\else
\language=\csname l@#1\endcsname
\fi
#2}}
\providecommand{\BIBdecl}{\relax}
\BIBdecl

\bibitem{correll2016analysis}
N.~Correll, K.~E. Bekris, D.~Berenson, O.~Brock, A.~Causo, K.~Hauser, K.~Okada,
  A.~Rodriguez, J.~M. Romano, and P.~R. Wurman, ``Analysis and observations
  from the first amazon picking challenge,'' \emph{IEEE Trans. Automation
  Science and Engineering}, vol.~15, no.~1, 2016.

\bibitem{morrison2018cartman}
D.~Morrison, A.~W. Tow, M.~Mctaggart, R.~Smith, N.~Kelly-Boxall, S.~Wade-Mccue,
  J.~Erskine, R.~Grinover, A.~Gurman, T.~Hunn \emph{et~al.}, ``Cartman: The
  low-cost cartesian manipulator that won the amazon robotics challenge,'' in
  \emph{ICRA}.\hskip 1em plus 0.5em minus 0.4em\relax IEEE, 2018, pp.
  7757--7764.

\bibitem{zeng2018robotic}
A.~Zeng, S.~Song, K.-T. Yu, E.~Donlon, F.~R. Hogan, M.~Bauza, D.~Ma, O.~Taylor,
  M.~Liu, E.~Romo \emph{et~al.}, ``Robotic pick-and-place of novel objects in
  clutter with multi-affordance grasping and cross-domain image matching,'' in
  \emph{ICRA}.\hskip 1em plus 0.5em minus 0.4em\relax IEEE, 2018, pp.
  3750--3757.

\bibitem{wan2020planning}
W.~Wan, K.~Harada, and F.~Kanehiro, ``Planning grasps with suction cups and
  parallel grippers using superimposed segmentation of object meshes,''
  \emph{IEEE Transactions on Robotics}, 2020.

\bibitem{mahler2018dex}
J.~Mahler, M.~Matl, X.~Liu, A.~Li, D.~Gealy, and K.~Goldberg, ``Dex-net 3.0:
  Computing robust vacuum suction grasp targets in point clouds using a new
  analytic model and deep learning,'' in \emph{ICRA}.\hskip 1em plus 0.5em
  minus 0.4em\relax IEEE, 2018.

\bibitem{Pham2019}
H.~Pham and Q.~C. Pham, ``{Critically fast pick-and-place with suction cups},''
  \emph{ICRA}, vol. 2019-May, pp. 3045--3051, 2019.

\bibitem{cheng2019manipulation}
X.~Cheng and M.~T. Mason, ``Manipulation with suction cups using external
  contacts,'' in \emph{ISRR}, 2019.

\bibitem{aoyagi2020bellows}
S.~Aoyagi, M.~Suzuki, T.~Morita, T.~Takahashi, and H.~Takise, ``Bellows suction
  cup equipped with force sensing ability by direct coating thin-film resistor
  for vacuum type robotic hand,'' \emph{IEEE/ASME Transactions on
  Mechatronics}, vol.~25, no.~5, pp. 2501--2512, 2020.

\bibitem{doi2020novel}
S.~Doi, H.~Koga, T.~Seki, and Y.~Okuno, ``Novel proximity sensor for realizing
  tactile sense in suction cups,'' in \emph{ICRA}.\hskip 1em plus 0.5em minus
  0.4em\relax IEEE, 2020.

\bibitem{eppner2016lessons}
C.~Eppner, S.~H{\"o}fer, R.~Jonschkowski, R.~Mart{\'\i}n-Mart{\'\i}n,
  A.~Sieverling, V.~Wall, and O.~Brock, ``Lessons from the amazon picking
  challenge: Four aspects of building robotic systems.'' in \emph{RSS}, 2016.

\bibitem{nadeau2020tactile}
P.~Nadeau, M.~Abbott, D.~Melville, and H.~S. Stuart, ``Tactile sensing based on
  fingertip suction flow for submerged dexterous manipulation,'' in
  \emph{ICRA}.\hskip 1em plus 0.5em minus 0.4em\relax IEEE, 2020, pp.
  3701--3707.

\bibitem{lederman1993extracting}
S.~J. Lederman and R.~L. Klatzky, ``Extracting object properties through haptic
  exploration,'' \emph{Acta psychologica}, vol.~84, no.~1, 1993.

\bibitem{li2020milliscale}
M.~S. Li, D.~Melville, E.~Chung, and H.~S. Stuart, ``Milliscale features
  increase friction of soft skin in lubricated contact,'' \emph{IEEE Robotics
  and Automation Letters}, vol.~5, no.~3, pp. 4781--4787, 2020.

\bibitem{shi1994good}
J.~Shi \emph{et~al.}, ``Good features to track,'' in \emph{Conference on
  computer vision and pattern recognition}.\hskip 1em plus 0.5em minus
  0.4em\relax IEEE, 1994, pp. 593--600.

\bibitem{hochreiter1997long}
S.~Hochreiter and J.~Schmidhuber, ``Long short-term memory,'' \emph{Neural
  computation}, vol.~9, no.~8, pp. 1735--1780, 1997.

\bibitem{chen2016xgboost}
T.~Chen and C.~Guestrin, ``Xgboost: A scalable tree boosting system,'' in
  \emph{22nd ACM SIGKDD Int. Conf. on knowledge discovery and data mining},
  2016.

\end{thebibliography}
\vspace{\baselineskip}

%% For Writing course editing

\end{document}